\documentclass[sigconf]{acmart}
\usepackage{multirow}
\usepackage{subfig}
\usepackage{color}
\usepackage{tikz}
\usepackage{graphicx}

\AtBeginDocument{%
  \providecommand\BibTeX{{%
    \normalfont B\kern-0.5em{\scshape i\kern-0.25em b}\kern-0.8em\TeX}}}

\copyrightyear{2024}
\acmYear{2024}
\setcopyright{acmlicensed}
\acmConference[MM '24] {Proceedings of the 32nd ACM International Conference on Multimedia}{October 28--November 1, 2024}{Melbourne, VIC, Australia.}
\acmBooktitle{Proceedings of the 32nd ACM International Conference on Multimedia (MM '24), October 28--November 1, 2024, Melbourne, VIC, Australia}
\acmISBN{979-8-4007-0686-8/24/10}
\acmDOI{10.1145/3664647.3680833}

\begin{document}\sloppy

\title{VrdONE: One-stage Video Visual Relation Detection}


\author{Xinjie Jiang}
\authornote{Equal contribution.}
\orcid{0009-0009-9207-9070}
\email{jiangxinjie512@gmail.com}
\author{Chenxi Zheng}
\orcid{0009-0006-0344-2439}
\authornotemark[1]
\email{chansey0529@gmail.com}
\affiliation{%
	\institution{South China University of Technology}
	\city{Guangzhou}
	\state{Guangdong}
	\country{China}
}

\author{Xuemiao Xu}
\authornote{Corresponding authors.}
\orcid{0000-0002-8006-3663}
\email{xuemx@scut.edu.cn}
\affiliation{%
	\institution{South China University of Technology}
	\city{Guangzhou}
	\state{Guangdong}
	\country{China}
}
\affiliation{%
	\institution{Guangdong Engineering Center for Large Model and GenAI Technology}
	\city{Guangzhou}
	\state{Guangdong}
	\country{China}
}

\author{Bangzhen Liu}
\authornotemark[2]
\orcid{0000-0001-6621-0594}
\email{liubz.scut@gmail.com}
\affiliation{%
	\institution{South China University of Technology}
	\city{Guangzhou}
	\state{Guangdong}
	\country{China}
}
\affiliation{%
	\institution{State Key Laboratory of Subtropical Building and Urban Science}
	\city{Guangzhou}
	\state{Guangdong}
	\country{China}
}

\author{Weiying Zheng}
\orcid{0000-0002-8196-1051}
\email{arisezheng21@gmail.com}
\affiliation{%
	\institution{South China University of Technology}
	\city{Guangzhou}
	\state{Guangdong}
	\country{China}
}

\author{Huaidong Zhang}
\orcid{0000-0001-7662-9831}
\email{huaidongz@scut.edu.cn}
\affiliation{%
	\institution{South China University of Technology}
	\city{Guangzhou}
	\state{Guangdong}
	\country{China}
}

\author{Shengfeng He}
\orcid{0000-0002-3802-4644}
\email{shengfenghe@smu.edu.sg}
\affiliation{%
	\institution{Singapore Management University}
	\city{Singapore}
	\state{}
	\country{Singapore}
}


\renewcommand{\shortauthors}{Xinjie Jiang et al.}

\newcommand{\rstred}[1]{{\textcolor[RGB]{210,21,19}{#1}}}
\newcommand{\rstgreen}[1]{{\textcolor[RGB]{22,112,8}{#1}}}

\begin{abstract}
\label{sec:abs}
Video Visual Relation Detection (VidVRD) focuses on understanding how entities interact over time and space in videos, a key step for gaining deeper insights into video scenes beyond basic visual tasks. Traditional methods for VidVRD, challenged by its complexity, typically split the task into two parts: one for identifying what relation categories are present and another for determining their temporal boundaries. This split overlooks the inherent connection between these elements. Addressing the need to recognize entity pairs' spatiotemporal interactions across a range of durations, we propose VrdONE, a streamlined yet efficacious one-stage model. VrdONE combines the features of subjects and objects, turning predicate detection into 1D instance segmentation on their combined representations. This setup allows for both relation category identification and binary mask generation in one go, eliminating the need for extra steps like proposal generation or post-processing. VrdONE facilitates the interaction of features across various frames, adeptly capturing both short-lived and enduring relations. Additionally, we introduce the Subject-Object Synergy (SOS) module, enhancing how subjects and objects perceive each other before combining. VrdONE achieves state-of-the-art performances on the VidOR benchmark and ImageNet-VidVRD, showcasing its superior capability in discerning relations across different temporal scales. The code is available at \textcolor[RGB]{228,58,136}{\href{https://github.com/lucaspk512/vrdone}{https://github.com/lucaspk512/vrdone}}.
\end{abstract}




\begin{teaserfigure}
	\includegraphics[width=\textwidth]{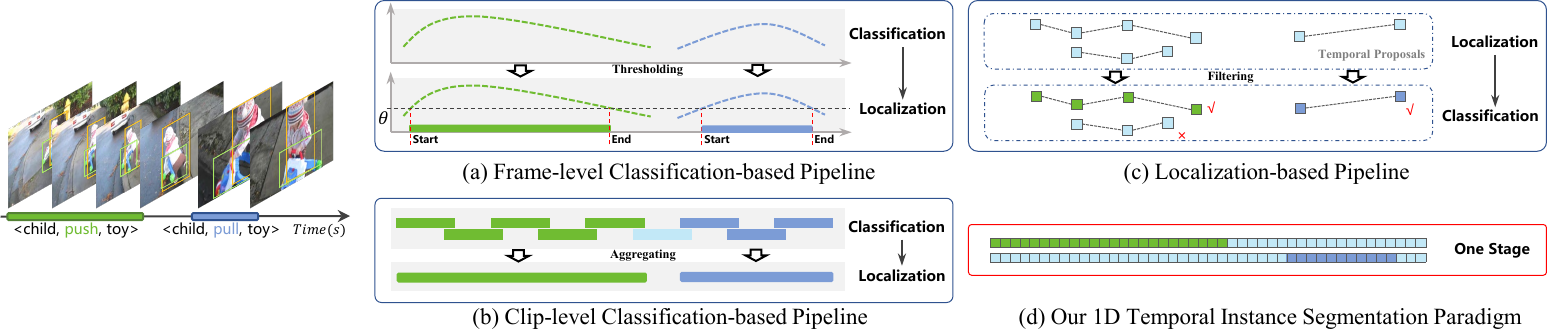}
	\vspace{-6mm}
	\caption{Classical pipelines for VidVRD include: (a) frame-level classification-based, (b) clip-level classification-based, and (c) localization-based approaches. These methods typically overlook the spatiotemporal interactions between entities. In contrast, our approach (d) utilizes a 1D temporal instance segmentation formulation that concurrently facilitates relation classification and per-frame relation mask generation for all relations in a single step, eliminating the need for additional post-processing.}
	\label{fig:fig_sec1_intro_teaser_formulation}
\end{teaserfigure}


\maketitle

\section{Introduction}
\label{sec:intro}

Deep learning has propelled significant enhancements in video visual analysis for a variety of tasks such as object tracking~\cite{meinhardt2022trackformer,cheng2022xmem,huang2017egocentric,huang2016stereo,he2015oriented}, action classification~\cite{feichtenhofer2019slowfast, wang2021actionclip,xu2020transductive}, and action localization~\cite{feichtenhofer2019slowfast, tong2022videomae, wang2023videomae}. Despite the advancements, the increasing complexity of video data requires precise interpretation of spatial and temporal relationships among entities in videos. To address this challenge, Video Visual Relation Detection (VidVRD) has been introduced. VidVRD aims to detect all relational instances in a video. Together with a pair of entities connected by the relation, each instance can be represented by a triplet $\langle \textit{subject, predicate, object} \rangle$. By harnessing rich semantic insights and interpretability, VidVRD is poised to enhance various downstream applications, including video captioning~\cite{yan2022videococa}, video question answering~\cite{yan2022videococa}, and video visual grounding~\cite{lin2023univtg}.

\begin{figure}[t]
    \centering
    \includegraphics[width=\linewidth]{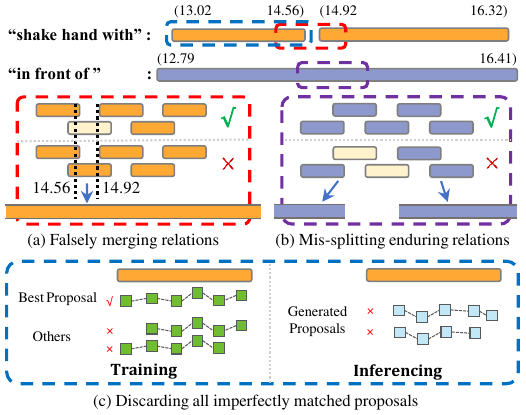}
    \vspace{-6mm}
    \caption{
        Limitations of existing two-stage methods. The heuristic aggregation in classification-based methods can lead to incorrect temporal localizations, causing (a) consecutive relations to be mistakenly identified as a single relation and (b) long-lasting relations improperly split into shorter segments. Localization-based methods also have drawbacks, where (c) relations might go undetected during inference due to mismatches with the fixed-length proposals.
    }
    \label{fig:fig_sec1_intro_teaser_failurecase}
\end{figure}

The VidVRD framework is divided into three sub-tasks: entity tracking, relation classification, and temporal boundary localization. As illustrated in Fig.~\ref{fig:fig_sec1_intro_teaser_formulation}, the process begins with the identification of each entity's category and spatial tracklets using pretrained video tracking models~\cite{chen2020memory,wojke2017simple,liu2019deformable}. Traditional approaches to VidVRD typically treat the tasks of classification and temporal localization as distinct, processing them sequentially in either a classification-based or localization-based manner. In classification-based strategies~\cite{shang2021video,zheng2022vrdformer}, relations are first identified on a clip-level~(Fig.~\ref{fig:fig_sec1_intro_teaser_formulation}(a)) or frame-level~(Fig.~\ref{fig:fig_sec1_intro_teaser_formulation}(b)), and then relation periods are determined using heuristic temporal aggregation algorithms~\cite{shang2021video, chen2021social}. Conversely, localization-based approaches~\cite{liu2020beyond}~(Fig.~\ref{fig:fig_sec1_intro_teaser_formulation}(c)) start with generating temporal proposals, which are then refined through a redundancy filtering mechanism before classification.

\begin{figure}[t]
    \centering
    \includegraphics[width=1\linewidth]{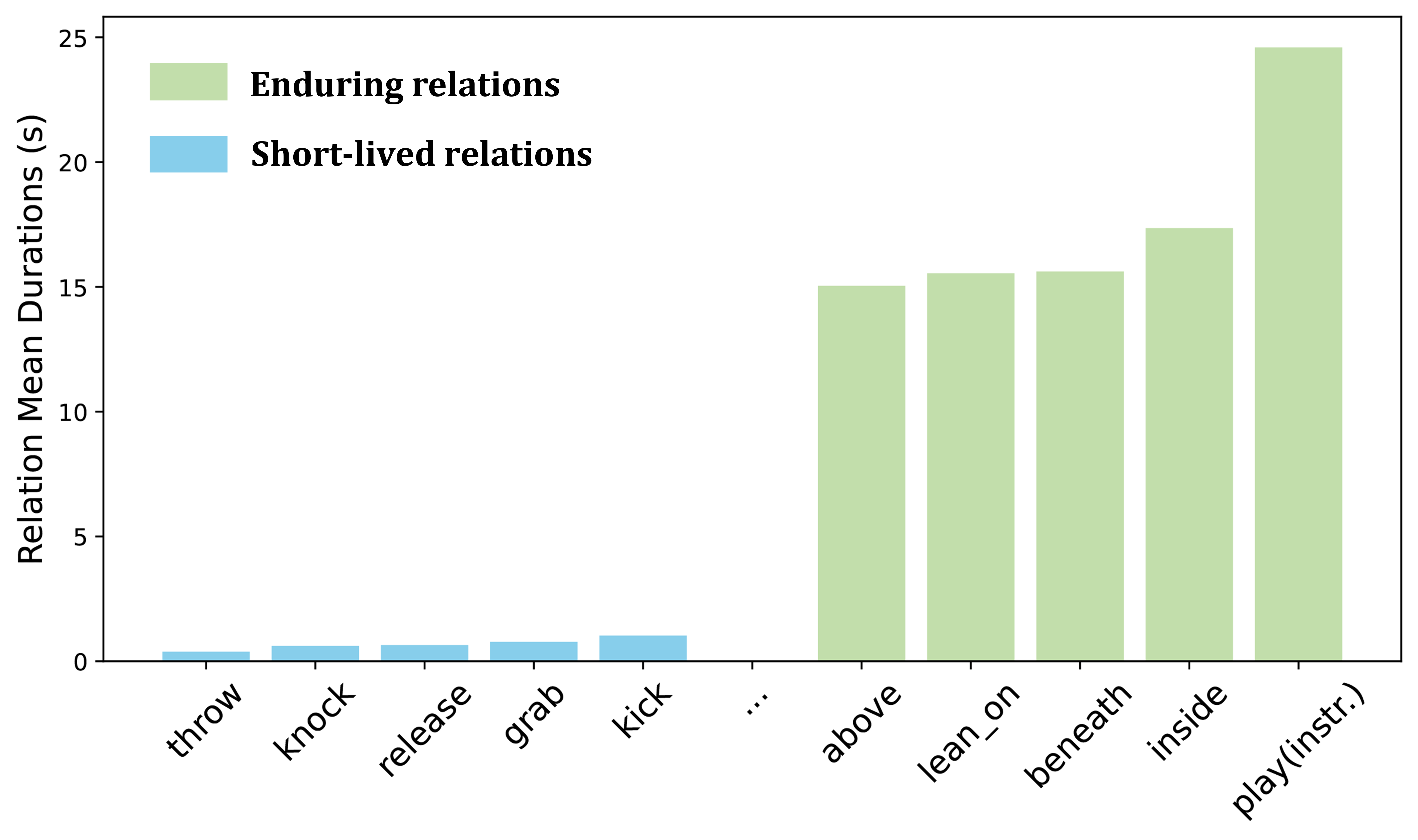}
    \vspace{-8mm}
    \caption{The mean duration distribution of the top 5 head and tail relations in the VidOR dataset, showing significant variation in their lengths.}
    \label{fig:fig_sec1_intro_teaser_distribution}
\end{figure}

However, existing methods do not coherently account for the spatiotemporal interactions between entities, resulting in suboptimal performance in both relation classification and localization. On the one hand, the integration of clip-level and frame-level short-term relations primarily depends on locally extracted features. This can lead to ambiguous detections at the temporal boundaries of relations, such as improperly merging temporally adjacent relations of the same category (Fig.~\ref{fig:fig_sec1_intro_teaser_failurecase}(a)) or mistakenly splitting a continuous relation into two disjoint ones (Fig.~\ref{fig:fig_sec1_intro_teaser_failurecase}(b)). On the other hand, generating proposals creates fixed-length temporal templates for video relations. As depicted in Fig.~\ref{fig:fig_sec1_intro_teaser_failurecase}(c), these templates often overlook potential relations that do not perfectly align with them during the inference stage, thereby constraining their effectiveness.

In real-world scenarios, interactions between entity pairs exhibit varied patterns across temporal and spatial dimensions. As shown in Fig.~\ref{fig:fig_sec1_intro_teaser_distribution}, the mean durations of relations vary significantly, with short-lived relations being transient while enduring relations can last dozens of times longer. Furthermore, entities within these relations differ in aspects such as movement speed and range. For instance, an enduring relation like ``in front of'' and a short-lived relation like ``shake hand with'' might co-occur between two individuals during the same video segment. While ``in front of'' might persist throughout the segment, ``shake hand with'' typically lasts only a few seconds and involves rapid hand movements.

This diversity in spatiotemporal dynamics underscores the importance of accurately accounting for these variations to categorize relation types. Motivated by these observations, we aim to enhance our model's performance by integrating richer spatiotemporal information in a local-to-global manner. Moreover, we unify video relation classification and temporal boundary localization into a single holistic problem, reformulating it as a 1D temporal instance segmentation task (see Fig.~\ref{fig:fig_sec1_intro_teaser_formulation}(d)). This unified approach allows for more precise relation classification and detailed relation boundary localization within a single inferencing step, benefiting from spatiotemporal synergistic learning and improved supervision provided by temporal location binary masks.

In this context, we introduce VrdONE, a spatiotemporal synergistic transformer designed for one-stage video visual relation detection. This model effectively detects all relation instances between subject-object pairs in an untrimmed video. Initially, we capture the temporal and spatial features of all entities in the video sequence. For each subject-object pair, we have developed the Subject-Object Synergy (SOS) module to improve their mutual perception. Additionally, a Bilateral Spatiotemporal Aggregation (BSA) mechanism has been designed to effectively learn features that encapsulate bilateral awareness for short-lived features of each other. Subject and object features are then fused and processed by a relation encoder to capture enduring relations. Finally, the unitied features are decoded by the classification and segmentation branches to obtain the relation categories and temporal boundaries simultaneously. Both branches are concurrently trained in a single stage.

In summary, our contributions are threefold:
\begin{itemize}
    \item We offer a novel perspective on the VidVRD challenge by reformulating it as a 1D instance segmentation task. This innovative approach allows simultaneous predicate category identification and binary mask generation for video relations in a single processing step.
    \item We propose VrdONE, a unique one-stage framework for VidVRD. Using Bilateral Spatiotemporal Aggregation, VrdONE enhances the interaction between subjects and objects across time and space, benefiting the detection of short-lived and enduring relations.
    \item Our experimental results on various benchmarks confirm that VrdONE sets a new standard for VidVRD. It significantly improves upon the state-of-the-arts in the VidVRD task, considering the performance on both the relation classification and temporal boundary localization sub-tasks.
\end{itemize}

\section{Related Work}
\label{sec:related_work}
\noindent{\textbf{Video Visual Relation Detection.}}
Recent advancements in VidVRD primarily fall into two categories: classification-based and localization-based methods. Utilizing features from pretrained tracking models~\cite{ren2015faster,danelljan2014accurate}, Shang~\etal~\cite{shang2017video} developped the first classification-based pipeline. This approach segments videos into short overlapping clips to classify relations within local ranges and employs an association algorithm for temporal localization based on adjacent classification results. Subsequent studies have refined this method by enhancing classification accuracy using graph convolution networks~\cite{qian2019video, wei2023defense} or integrating multi-modal features~\cite{wei2023defense, su2020video}. However, clip-based approaches struggle with prolonged relations and are prone to errors from cumulative association steps, despite some algorithms~\cite{su2020video,wei2023defense} being proposed to mitigate the problem. To better capture long-range relations, Chen~\etal~\cite{chen2021social} introduced a multi-modal prototype learning approach for frame-level classification, with a 1D watershed algorithm~\cite{roerdink2000watershed} for temporal localization. Concurrently, Gao~\etal~\cite{gao2022classification} explored a two-stage learning strategy for classification and localization separately. Zheng~\etal~\cite{zheng2022vrdformer} proposed learning entity tracklets and their relation categories simultaneously to mitigate errors in generating entity tracklets, while still localizing temporal boundaries with heuristic aggregation algorithms. Contrarily, Liu~\etal~\cite{liu2020beyond} attempted a localization-based approach by generating numerous temporal proposals through sliding windows, which were then filtered through template matching to accurately determine relation durations. Differing from these approaches, we reconceptualize the challenges of classification and temporal localization into a one-stage framework to capture their inherent connection. Our method leverages interactions between subject and object features across frames to effectively capture short-lived and enduring relations, significantly improving the two sub-tasks.

\begin{figure*}[tp]
    \centering
    \includegraphics[width=\textwidth]{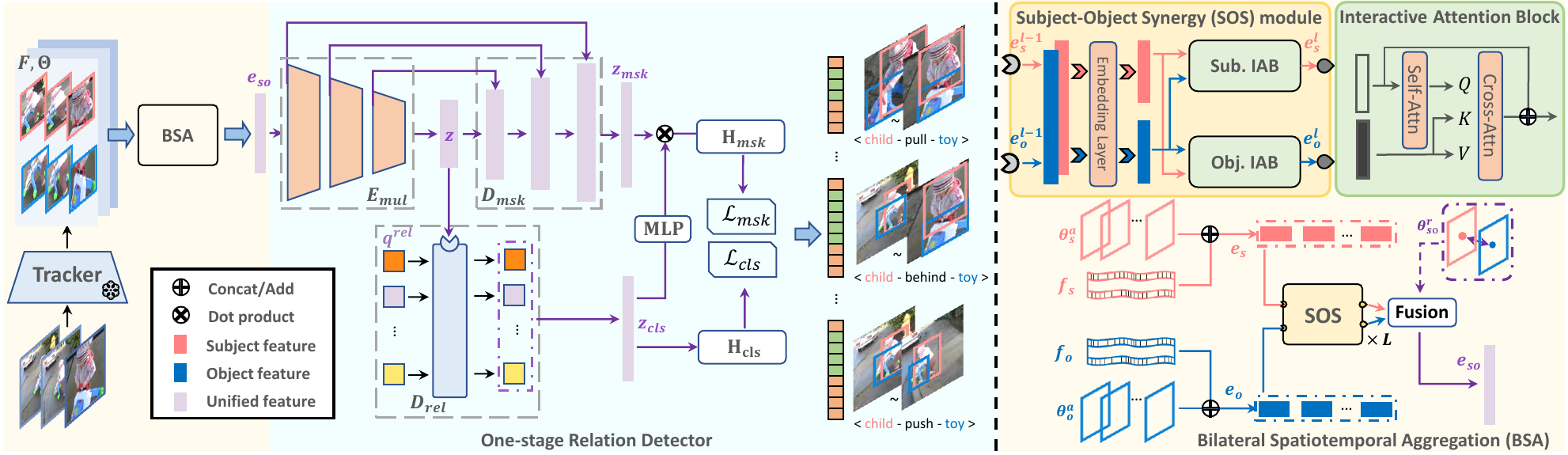}
    \vspace{-6mm}
    \caption{
      The VrdONE pipeline processes an untrimmed video by first extracting visual and spatial change features ($f$ and $\theta$) for all entities' tracklets using a frozen pretrained video tracker. For each subject-object pair, the BSA then encapsulates bilateral awareness into the feature embeddings. Absolute positional changes ($\theta^a$) are injected into entity features, followed by $L$ SOS modules to enrich spatiotemporal interactions. After equipping these enriched embeddings with relative spatial features $\theta^{r}$, the resulting unified embeddings $e_{so}$ are further processed by the relation encoder $E_{mul}$ and directed to two synergistic decoder $D_{msk}$ and $D_{rel}$. With the help of the generated temporal-aware features $z_{msk}$ and category-aware features $z_{cls}$, VrdONE achieves one-stage processing for both video relation classification and temporal localization.
    }
    \label{fig:fig_sec3_method_pipeline}
\end{figure*}

\noindent{\textbf{Spatiotemporal Synergistic Learning in Videos.}}
Understanding vision tasks in videos requires a spatiotemporal synergistic approach. Initially, 3D convolutional neural networks were used to extract features across both spatial and temporal dimensions~\cite{carreira2017quo, feichtenhofer2019slowfast}. More recently, transformer architectures have brought significant advancements in computer vision~\cite{vaswani2017attention, dosovitskiy2020image, liu2021swin}. For instance, ViViT~\cite{arnab2021vivit} integrates these architectures into video processing and sets new performance benchmarks, surpassing older 3D convolution-based methods. The Video Swin Transformer~\cite{liu2022video} adapts the Swin Transformer~\cite{liu2021swin} concept to video by expanding it into three dimensions, which enhances information capture from local to global contexts, improving learning efficiency. Similarly, VideoMAE~\cite{tong2022videomae} and its successor, VideoMAE V2~\cite{wang2023videomae}, leverage a Masked AutoEncoder~\cite{he2022masked} approach in a self-supervised learning framework, applying consistent spatial masks across video snippets to increase model robustness and effectiveness, thereby achieving notable performance improvements in various video processing tasks. Integrating spatiotemporal elements is crucial for optimizing video processing across diverse applications.

\section{Methods}
\label{sec:method}

\subsection{Preliminaries}

\noindent{\textbf{Problem Setting.}} Given an untrimmed video $V$ that contains $N$ entities (including subjects and objects), the goal of VidVRD is to learn a video relation detector $\mathcal{G}$ that generates all possible relations $R$ and their duration ranges $T_b, T_e$ between entity pairs. Here, $T_b$ and $T_e$ are the begin and end frame indices of relations. Concretely, the detection results are formulated as:
\begin{equation}
    \mathcal{G}(F) = \{(\langle s_i, r, o_j \rangle, t_{b}, t_{e})\}, i, j \in[1, N],
    \label{eq:setting}
\end{equation}
where $s_i$ and $o_j$ denote the subject and object categories and the corresponding spatial location tubelets. In this case, $F = \{ f_1, f_2, ..., f_N \}$ represents the extracted features of entities. The tracklet features for entity $i$ are $f_i \in \mathbb{R}^{l_i \times C}$, a set of feature vectors with dimension $C$ extracted from uniformly sampled consecutive video frames. Here, $l_i$ is the duration length of entity $i$.

Typically, the pipeline of VidVRD can be divided into three sub-tasks: entity tracking, relation classification, and temporal boundary localization. After extracting $F$ using an entity detector and tracker, previous works often treat relation classification and temporal boundary localization separately. This procedure can be explained by Bayes's Formula. Concretely, the distribution of detection results is formulated either as $P(R, T_b, T_e | F) = P(T_b, T_e | R, F) P(R | F)$ or $P(R, T_b, T_e | F) = P(R | T_b, T_e, F) P(T_b, T_e | F)$, which are the common practices for classification-based and localization-based methods, respectively. However, ignoring the inherent connection between the two sub-tasks deteriorates both performances. To fully exploit the temporal and spatial features during the interaction of subjects and objects, we propose to reformulate the problem in a one-stage manner, \ie, directly estimating $P(R, T_b, T_e | F)$.

\noindent{\textbf{Local Attention Mechanism.}}
The attention mechanism~\cite{vaswani2017attention, dosovitskiy2020image} has demonstrated its great ability to capture global information along an input sequence. Given the input query $q$, key $k$, and value $v$, the attention operation with multiple heads $\mathcal{H}$ is calculated as:
\begin{equation}
    \begin{aligned}
        \mathbf{MA}(q, k, v) &= \mathbf{Concat}(\mathcal{H}_1, ..., \mathcal{H}_H)W^o, \\
        \mathcal{H}_{i} &= \mathbf{Softmax}(\frac{qW_i^q(kW_i^k)}{\sqrt{d_k}})vW_i^v, 
    \end{aligned}
    \label{eq:attention}
\end{equation}
where $H$ is the number of heads, $W_i^q, W_i^k, W_i^v, W^o$ are projection operations, and $d_k$ is the dimension of the key.

As shown in Fig.~\ref{fig:fig_sec1_intro_teaser_distribution}, the mean durations across different types of relations vary drastically, indicating the necessity to capture dynamic changes of entities from local to global. Additionally, temporal context beyond a certain range is not effective for temporal localization~\cite{zhang2022actionformer}. To capture local information within the neighboring region of the input sequence, we use local attention~\cite{choromanski2020rethinking} to restrict the attention scopes on the feature sequence. When the network goes deeper, receptive fields of stacked local attention layers will gradually extend to global ranges, similar to CNNs and the Swin Transformer~\cite{liu2021swin}, thus capturing both short-lived and enduring relations.

Concretely, the input $q, k$, and $v$ are all divided into $M$ overlapping chunks with a window size of $k_w$. Consequently, the local attention calculation is performed as the concatenation of attention results in chunks:
\begin{align}
    \mathbf{LocalMA}(q, k, v) = \mathbf{Concat}(\mathbf{MA}(q_1, k_1, v_1), ..., \mathbf{MA}(q_M, k_M, v_M)).
    \nonumber
\end{align}
To more conveniently demonstrate our VrdONE in the following sections, we define the calculation of local self-attention ($\mathbf{LocalMSA}$) and local cross-attention ($\mathbf{LocalMCA}$) as: 
\begin{align}
    \mathbf{LocalMSA}(e) &= \mathbf{LocalMA}(e, e, e), \label{eq:localmsa}\\
    \mathbf{LocalMCA}(e_i, e_j) &= \mathbf{LocalMA}(e_i, e_j, e_j). \label{eq:localmca}
\end{align}
The calculations of vanilla global self-attention $\mathbf{MSA}(e)$ and cross-attention $\mathbf{MCA}(e_i, e_j)$ are similar to Eq.~\eqref{eq:localmsa} and Eq.~\eqref{eq:localmca}.

\subsection{Overview}
The overall pipeline of VrdONE is depicted in Fig.~\ref{fig:fig_sec3_method_pipeline}.  Firstly, we apply a fixed pretrained object tracker~\cite{ren2015faster,chen2020memory} to extract entities' features $F$. For each subject-object pair, we process their features using the Bilateral Spatiotemporal Aggregation (Sec.~\ref{sec:method_bsa}) to fully perceive spatiotemporal interactions in the video. Specifically, we propose a Subject-Object Synergy module to enhance the mutual perception between the two entities. The resulting unified embeddings further proceed to the one-stage relation detector (Sec.~\ref{sec:method_one_stage}) for classification and temporal localization. The one-stage relation detector consists of a multi-scale relation encoder, a relation decoder, and a temporal mask generator. As the receptive fields extend, our model can focus on both short-term and long-term changes in feature representations. The classification and segmentation branches are concurrently trained in a single stage by a relation identification loss and a mask prediction loss.

\subsection{Bilateral Spatiotemporal Aggregation}
\label{sec:method_bsa}

In Bilateral Spatiotemporal Aggregation (BSA), we promote bilateral awareness of subject and object features through mutual perception. Ultimately, a pair of features are encoded into a unified relational representation for subsequent one-stage instance segmentation.

Given the tracklet features $F$ of entities, we generate several subject-object pairs $(f_s, f_o)$, where $f_s, f_o \in F$. Since the visual temporal ranges of the subject and object typically differ, we retain only the features within the overlapping time range $l_{so}$ to obtain synchronized feature vectors, trimming the features as $f_s, f_o \in \mathbb{R}^{l_{so} \times C}$.

To incorporate spatial information detected by the former tracker into visual features, we adopt the approach from~\cite{gao2022classification} and employ absolute positional representations $\theta_s^{a}$ and $\theta_o^{a}$, where $\theta_s^{a}, \theta_o^{a} \in \mathbb{R}^{l_{so} \times 8}$. To be specific, these positional representations comprise the normalized bounding box coordinates and the corresponding offsets between two consecutive frames. Thereafter, the visual features $f$ and absolute spatial features $\theta^{a}$ are integrated into general entity embeddings using a multilayer perceptron (MLP):
\begin{align}
    e = \mathbf{MLP}(\mathbf{Concat}(f, \theta^a)).
  \label{eq:entity_fuse}
\end{align}
We omit the subscripts here without introducing ambiguity. This process integrates the visual and spatiotemporal features into subject and object embeddings $e_s$ and $e_o$, which are then fed into the Subject-Object Synergy module to comprehend interactions. Positional changes between consecutive frames for individual entities provide action-related hints for VrdONE, making our method capable of capturing relations with various movement patterns. 

\noindent{\textbf{Subject-Object Synergy Module.}} The Subject-Object Synergy (SOS) module facilitates interaction between subject and object embeddings to enhance mutual understanding. It consists of an embedding layer and two Interactive Attention Blocks (IABs).

The embedding layer, sharing the same structure as the encoder layer in transformer~\cite{vaswani2017attention}, includes a local multi-head self-attention layer and an MLP. Specifically, given that there are totally $L$ SOS blocks, the embedding layer of the $l_{th}$ block is defined as:
\begin{equation}
    \begin{aligned}
        \bar{e}^{l} &= \mathbf{LocalMSA}(e^{l-1}) + e^{l-1}, \\
        \hat{e}^{l} &= \mathbf{MLP}(\bar{e}^{l}) + \bar{e}^{l}.
    \end{aligned}
    \label{eq:sos_embedding}    
\end{equation}
By applying Eq.~\eqref{eq:sos_embedding} to the subject and object inputs, \ie, $e_s^{l-1}$ and $e_o^{l-1}$, we obtain the embedded features $\hat{e}_s^l$ and $\hat{e}_o^l$ for the subject and object respectively.

Subsequently, we process the embeddings with the IABs. The IAB enables information exchanges between subject and object features to enrich their representations. Concretely, the IAB is composed of a local self-attention and cross-attention layer. For instance, to integrate object features into subject features, the aggregated representations for the subject are expressed as:
\begin{equation}
    \begin{aligned}
        \tilde{e}_s^{l} &= \mathbf{LocalMSA}(\hat{e}^{l}_s), \\
        e_s^{l} &= \mathbf{LocalMCA}((\tilde{e}^{l}_s, \hat{e}^{l}_o)) + \hat{e}_s^{l}. \\
    \end{aligned}
    \label{eq:sos_iab}    
\end{equation}
Likewise, we augment the object features with mutual information from the subject and finally obtain aggregated object features $e_o^{l}$. Early mutual perception between subject and object is crucial when the receptive field is relatively restricted, allowing attention to relation-relevant features while discarding irrelevant ones. Additionally, according to ablation studies, deeper layer interactions in SOS are found to be less effective.

\noindent{\textbf{Pair Embedding Fusion.}}
After applying $L$ SOS layers, the enhanced subject and object embedding features ($e_s^{L}$ and $e_o^{L}$) capture comprehensive representations with innovative features from their interactions. The receptive fields extended by stacked local attention layers can already cover short-lived relations. However, some relations, especially enduring ones, exhibit distinctive co-spatial patterns. For instance, in the case of ``in front of'', the absolute positions of individuals may change without rules, but their relative positions usually remain in a consistent direction.

This indicates that we need to treat the pair as a unity instead of two separate individuals in the later stage of our model. We subsequently fuse the subject and object features to form unified representations $e_{so}$. To further facilitate positional awareness for spatiotemporal synergism, we inject the relative positional features $\theta^r_{so} \in \mathbb{R}^{l_{so} \times 5}$~\cite{li2021interventional}, defined as:
\begin{equation}
    \begin{aligned}
        \theta^r_{so} = [\frac{x^s - x^o}{x^o}, \frac{y^s - y^o}{y^o}, \log \frac{w^s}{w^o}, \log \frac{h^s}{h^o}, \log \frac{w^s \cdot h^s}{w^o \cdot h^o}],
    \end{aligned}
    \label{eq:relative_bbox_feature}
\end{equation}
where $x^{s/o}$ and $y^{s/o}$ are the center's coordinates, and $w^{s/o}$ and $h^{s/o}$ represent the width and height of the bounding box, respectively. Specifically, $\theta^r_{so}$ represents the relative spatial changes between the subject and object, consisting of four components for the relative shifts of coordinates and one for the changes of relative areas. 

Finally, adhering to the minimal design principle, we fuse the pair and inject spatial features with the MLP. The five-dimensional vectors in $\theta_{so}^r$ are projected using a one-dimensional convolution to match the dimensionality with the embeddings before combining. The unified representations $e_{so}$ are calculated as:
\begin{align}
    \hat{e}_{so} &= \mathbf{MLP}(\mathbf{Concat}(e_s^{L}, e_o^{L})), \\
    e_{so} &= \mathbf{MLP}(\mathbf{Concat}(\hat{e}_{so}, \mathbf{Conv1D}(\theta^r_{so}))).
  \label{eq:entity_relation}
\end{align}

\subsection{One-stage Relation Detector}
\label{sec:method_one_stage}

After obtaining the unified embeddings $e_{so}$ that contain rich spatiotemporal information, we process it to achieve simultaneous relation classification and temporal localization through a one-stage relation detector. The detector comprises a relation encoder $E_{mul}$, a relation decoder $D_{rel}$, and a temporal mask decoder $D_{msk}$.   

We follow the design of the feature pyramid network~\cite{lin2017feature} for the relation encoder and 1D mask decoder. The $E_{mul}$ consists of multiple transformer encoder blocks, which share a similar architecture to that in Eq.~(\ref{eq:sos_embedding}). The relation encoder continues to expand the receptive field to perceive long-range temporal information, thus benefiting enduring relations. 1D max pooling operations are applied between transformer blocks to downsample and generate a low-resolution feature map $z \in \mathbb{R}^{(l_{so} / l_{std}) \times C}$ in the end, where $l_{std}$ is the stride. Then, $D_{msk}$ gradually upsamples $z$ to recover the resolution for fine-grained per-frame binary mask generation. Concretely, lower-resolution features in the feature pyramid are combined with higher-resolution features via lateral connections.

We employ a query-based transformer as our relation decoder. For relation classification, $D_{rel}$ takes $z$ as its input to access high-dimensional semantic information. Specifically, the relation decoder consists of stacked transformer decoder blocks with $N_q$ learnable query embeddings $q^{rel}\in \mathbb{R}^{N_q \times C}$, which serve as pattern learners for all possible relation instances within a pair. It is worth noting that we utilize vanilla global multi-head self-attention $\mathbf{MSA}$ and cross-attention $\mathbf{MCA}$ here instead of their local versions. The reason for not using local attention is to ensure that query tokens remain independent of each other, without introducing any positional correlations that could influence the decoded results. Consequently, the local mechanism is not unsuitable for extracting the query features, as the local attention incorporates information from adjacent query tokens. The outputs of the relation decoder, denoted as $z_{cls}$, are projected and then used to filter decoded mask features $z_{msk}$ via a dot product to generate binary masks for different relation instances. Finally, the classification and localization head ($H_{cls}$ and $H_{msk}$) compute the relation categories and binary masks.

\begin{table*}[t]
    \centering
    \caption{Comparison with state-of-the-arts on the VidOR dataset. For object detectors, ``FR'', ``MG'', and ``IE'' symbolize Faster R-CNN~\cite{ren2015faster}, MEGA~\cite{chen2020memory}, and Integrated Encoder, respectively. For extra features, ``L'' and ``M'' denote language and mask features, whereas I3D~\cite{carreira2017quo} and CLIP~\cite{radford2021learning} denote visual feature extractor. For Social Fabric and our VrdONE, we represent the variants with extra features with a ``-X'' postfix. The best and second-best performances are bolded and underlined.}
    \vspace{-3mm}
    \scalebox{1}{
        \begin{tabular}{c|c|c|ccc|ccc}
            \toprule
            \multirow{2}{*}{Method} & \multirow{2}{*}{Detector} & \multirow{2}{*}{\shortstack{Extra\\Feature}} & \multicolumn{3}{c|}{Relation Detection} & \multicolumn{3}{c}{Relation Tagging} \\
            & & & mAP & R@50 & R@100 & P@1 & P@5 & P@10 \\
            \midrule
            TSPN~\cite{woo2021and} & FR & -- & 7.61 & 9.33 & 10.71 & 53.14 & 42.22 & 34.94 \\
            VidVRD-II~\cite{shang2021video} & FR & -- & 8.65 & 8.59 & 10.69 & 57.40 & 44.54 & 33.30\\
            BIG~\cite{gao2022classification} & MG & I3D+L & 8.54 & 8.03 & 10.04 & 64.42 & 51.80 & 40.96 \\
            HCM~\cite{wei2023defense} & MG & -- & 10.44 & 9.74 & 11.23 & 67.43 & 52.19 & 40.30 \\
            VRDFormer~\cite{zheng2022vrdformer}  & IE  & -- & 11.19 & 11.05 & 13.34 & 63.71 & 51.07 & 39.89 \\
            Social Fabric~\cite{chen2021social} & FR & I3D & 9.54 & 8.49 & 10.17 & 59.24 & 47.24 & 35.99 \\
            Social Fabric-X~\cite{chen2021social} & FR & I3D+L+M & 11.21 & 9.99 & 11.94 & \textbf{68.86} & \underline{55.16} & 43.40 \\
            \midrule
            \textbf{VrdONE} & MG & -- & \underline{11.86} & \underline{11.13} & \underline{14.21} & 66.11 & 54.92 & \underline{43.90}\\
            \textbf{VrdONE-X} & MG & CLIP & \textbf{12.17} & \textbf{11.41} & \textbf{14.55} & \underline{67.67} & \textbf{55.58} & \textbf{44.28} \\
            \bottomrule
        \end{tabular}
    }
    \label{tab:vidor_metrics}
    \centering
\end{table*}

\subsection{Training and Inference}

\noindent{\textbf{Loss Functions.}}
Similar to MaskFormer \cite{cheng2021per}, we employ a bipartite matching strategy to assign different queries to learn the corresponding instances. The matching cost for relation identification and binary mask prediction is denoted as:  
\begin{align}
    \mathcal{L}_{match} = \lambda_{cls} \cdot \mathbf{CE}(\hat{p_i}, c_j^{gt}) 
            + \mathcal{L}_{mask}(\hat{m}_i, m_j^{gt}),
    \label{eq:matching_cost}
\end{align}
where $\lambda_{cls}$ is the weight for the classification cost and the cost $-\hat{p_i}(c_j^{gt})$ used in DETR~\cite{carion2020end} is replaced by the cross-entropy loss $\mathbf{CE}(\cdot, \cdot)$ for better performance. The mask cost $\mathcal{L}_{mask}$ is composed of a binary focal loss \cite{lin2017focal} $\mathbf{FL}(\cdot, \cdot)$ and a dice loss \cite{milletari2016v} $\mathbf{Dice}(\cdot, \cdot)$ multiplied by hyper-parameters $\lambda_{mf}$ and $\lambda_{md}$ respectively:
\begin{align}
    \mathcal{L}_{mask} = \lambda_{mf} \cdot \mathbf{FL}(\hat{m}_i, m_j^{gt}) + \lambda_{md} \cdot \mathbf{Dice}(\hat{m}_i, m_j^{gt}).
    \label{eq:mask_cost}
\end{align}
The overall loss function for training is given by:
\begin{align}
    \mathcal{L} = \lambda_{cls} \cdot \mathbf{CE}(\hat{p}_{\sigma(j)}, 
    c_j^{gt}) + \mathbb{I}_{c_j^{gt} \neq \varnothing} \mathcal{L}_{mask}(\hat{m}_{\sigma(j)}, m_j^{gt}),
\end{align}
where $\sigma(j)$ denotes the index of the query token matched to the ground truth with index $j$. 

\noindent{\textbf{Fusing Additional Features.}}
Previous methods~\cite{chen2021social,gao2022classification,liu2020beyond} employ multiple additional features~\cite{carreira2017quo,mikolov2013efficient} from pretrained models to provide hints to presented models. Although VrdONE can fully extract the interaction patterns between subjects and objects, we still observe improvements with additional visual features~\cite{radford2021learning}. Before BSA, additional features are combined with entity features by a simple method similar to Eq.~(\ref{eq:entity_fuse}).

\noindent{\textbf{Inference Phase.}}
Subject-object pairs are generated based on the annotations during training. In the inference phase, we exhaustively enumerate all possible pairs within the current video, resulting in $N\times(N-1)$ potential subject-object pairs for detection. However, VrdONE can detect relations for the pairs in parallel and output all results in one step. For segmented frames, we consider those with a foreground probability greater than $0.5$ as the detected relation range, with the boundaries of $t_b$ and $t_e$ being the indices of the first and last foreground frames. Any post-processing to remove isolated noisy positive points is ignored, as our model demonstrates significant robustness in accurately segmenting the positive frames.

\begin{table*}[t]
    \centering
	\caption{Comparison with state-of-the-arts on ImageNet-VidVRD dataset. $^\dag$ denotes the version implemented by the authors.}
    \vspace{-3mm}
    \scalebox{1}{
        \begin{tabular}{c|c|c|ccc|ccc}
            \toprule
            \multirow{2}{*}{Method} & \multirow{2}{*}{Detector} & \multirow{2}{*}{\shortstack{Extra\\Feature}} & \multicolumn{3}{c|}{Relation Detection} & \multicolumn{3}{c}{Relation Tagging} \\
            & & & mAP & R@50 & R@100 & P@1 & P@5 & P@10 \\
            \midrule
			VRD-STGC~\cite{liu2020beyond} & FR & I3D & 18.38 & 11.21 & 13.69 & 60.00 & 43.10 & 32.24\\
            Social Fabric~\cite{chen2021social} & FR & -- & 19.23 & 12.74 & 16.19 & 57.50 & 43.40 & 31.90 \\
            Social Fabric-X~\cite{chen2021social} & FR & I3D+L+M & 20.08 & 13.73 & 16.88 & 62.50 & 49.20 & 38.45 \\
            VidVRD-II$^\dag$~\cite{shang2021video} & FR & -- & 23.85 & 9.74 & 10.86 & 73.00 & 53.20 & 39.75\\
            BIG~\cite{gao2022classification} & MG & -- & 26.08 & 14.10 & 16.25 & 73.00 & 55.10 & 40.00 \\
            HCM~\cite{wei2023defense} & MG & -- & \underline{29.68} & \underline{17.97} & \underline{21.45} & \underline{78.50} & \underline{57.40} & \underline{43.55} \\
            \midrule
            \textbf{VrdONE} & MG & -- & \textbf{31.33} & \textbf{18.20} & \textbf{21.61} & \textbf{80.50} & \textbf{59.40} & \textbf{44.17} \\
            \bottomrule
        \end{tabular}
    }
    \label{tab:vidvrd_metrics}
    \centering
\end{table*}

\begin{table}[tp]
    \caption{Ablation on the SOS module. ``w/o SOS'' denotes the removal of SOS. ``w/o IAB'', ``Cross'' and ``IAB'' indicate SOS with the removal of the IAB module, basic cross-attention, and IAB, respectively. $^*$ indicates our implementation.}
    \vspace{-3mm}
    \centering
    \scalebox{0.9}{
        \begin{tabular}{c|ccc|ccc}
            \toprule
            \multirow{2}{*}{Approach} &  \multicolumn{3}{c|}{Relation Detection} & \multicolumn{3}{c}{Relation Tagging} \\
            & mAP & R@50 & R@100 & P@1 & P@5 & P@10 \\
            \midrule
            w/o SOS & 11.28 & 10.83 & 13.64 & 65.74 & 54.68 & \textbf{44.06} \\
            \cline{1-7}
            w/o IAB & 11.60 & 10.97 & 14.01 & 65.98 & 54.54 & 43.79 \\
            Cross & \underline{11.72} & \underline{11.09} & \underline{14.11} & \textbf{66.82} & \underline{54.87} & 43.74 \\
            IAB$^*$ & \textbf{11.86} & \textbf{11.13} & \textbf{14.21} & \underline{66.11} & \textbf{54.92} & \underline{43.90} \\
            \bottomrule
        \end{tabular}
    }
    \label{tab:vidor_mcqa_arch_ablation}
    \centering
\end{table}

\begin{table}[tp]
    \caption{Parameter sensitivity analysis for the number of queries $N_q$.}
    \vspace{-3mm}
    \centering
    \scalebox{0.9}{
        \begin{tabular}{l|ccc|ccc}
            \toprule
            \multirow{2}{*}{$N_q$} &  \multicolumn{3}{c|}{Relation Detection} & \multicolumn{3}{c}{Relation Tagging} \\
            & mAP & R@50 & R@100 & P@1 & P@5 & P@10 \\
            \midrule
            5 & 11.62 & 11.02 & 13.98 & \underline{66.59} & 54.06 & 43.16 \\ 
            7 & \underline{11.82} & \underline{11.08} & 14.10 & 66.23 & \textbf{55.17} & \underline{43.93} \\
            9$^*$ & \textbf{11.86} & \textbf{11.13} & \textbf{14.21} & 66.11 & \underline{54.92} & 43.90 \\
            11 & 11.66 & 11.00 & 13.98 & 66.11 & 54.85 & 43.89 \\
            13 & 11.59 & 11.03 & \underline{14.16} & \textbf{66.95} & 54.47 & \textbf{43.97} \\
            \bottomrule
        \end{tabular}
    }
    \label{tab:vidor_query_ablation}
    \centering
\end{table}

\section{Experiments}
\label{sec:exp}

\noindent{\textbf{Datasets.}}
We evaluate our VrdONE on two VidVRD datasets: VidOR~\cite{shang2019annotating} and ImageNet-VidVRD~\cite{shang2017video}. The VidOR is a challenging dataset with a total duration of approximately 98.6 hours. It includes a training set with 7,000 videos, a validation set with 835 videos, and a test set with 2,165 videos. There are 80 entity categories and 50 predicate categories in the dataset. VidOR provides densely annotated entity bounding boxes in each frame and the detailed temporal boundaries of frame indices for visual relations between subject-object pairs. Due to missing labels in the test set, we train our method on the training set and evaluate the validation set following common practice. ImageNet-VidVRD is much smaller (about 3 hours) and easier, with only 1,000 videos, 800 for training and 200 for testing. It contains 35 and 132 categories for entity and predicate. Temporal boundaries of relations in ImageNet-VidVRD are labeled coarsely with relation lengths as multiples of 15 frames.

\noindent{\textbf{Evaluation Metrics.}}
We assess VrdONE's performance in relation detection (RelDet) and relation tagging (RelTag). (1) RelDet is a comprehensive task for both evaluating entity tracklets and temporal detection of relations. A detected triplet is correct if the triplet classes match the ground truth, and the entity tracklets and relation temporal boundaries sufficiently overlap with the ground truth, \eg, $vIoU>0.5$ and $tIoU > 0.5$. We utilize mAP and top K Recalls (R@K, where K=50, 100) as metrics. (2) RelTag is a simpler task, solely judging if the top K classification results of triplets happen within the videos. It disregards the localization results of both tracklets and relations. Top K precisions (P@K, where K=1, 5, 10) are employed as the evaluation metrics for RelTag.

\noindent{\textbf{Implementation Details.}}
Following~\cite{gao2022classification,wei2023defense}, we utilize the pretrained Video Object Detector MEGA~\cite{chen2020memory} to detect entities and extract their visual features, providing $F$ and $\Theta$ (\ie, $\Theta^a$ and $\Theta^r$) our model needs. Detection results are consolidated into object tracklets using deepSORT~\cite{wojke2017simple}. We set the maximum length of overlapped subject-object durations $l_{so}$ to 512, otherwise cutting out the outer length. The number of SOS block $L$ is 2. $E_{mul}$ contains 3 blocks, alongside the output $e_{so}$ from BSA, resulting in a 4-layer feature pyramid. With a downsampling ratio of 2, the feature pyramid comprises lengths of $[512, 256, 128, 64]$, and the downsampling ratio $l_{std}$ is 8. We set $k_w$ to 9. The decoder consists of 4 layers, with the number of queries $N_q$ set to 9. Parameters $\lambda_{cls}, \lambda_{mf}, \lambda_{md}$ are set to 2, 2, and 5, respectively. Prior to each local attention and MLP computation, LayerNorm~\cite{ba2016layer} is added. Drop-path~\cite{larsson2016fractalnet} is applied with a ratio of $0.1$. Training of VrdONE employs the AdamW~\cite{loshchilov2017decoupled} optimizer with a learning rate of $2\times10^{-4}$. Warmup and Exponential Moving Average (EMA) techniques are employed to enhance and stabilize the training process.

\subsection{Comparison with State-of-the-Arts}
We conduct experiments on the VidOR and ImageNet-VidVRD datasets and compare our VrdONE with the state-of-the-art methods on RelDet and RelTag tasks, as illustrated in Table~\ref{tab:vidor_metrics} and Table~\ref{tab:vidvrd_metrics}.

On the VidOR dataset, we implement two versions of VrdONE: a base model with only features provided by the video object detector, and an improved version, noted as VrdONE-X, incorporating visual features extracted by the CLIP~\cite{radford2021learning} image encoder. Even in its base implementation, VrdONE achieves state-of-the-art performance on most metrics. In particular, VrdONE exhibits noticeable improvements (+0.65\% mAP, +0.08\% R@50, and +0.87\% R@100) on all the RelDet metrics compared to the previous best performance, indicating a comprehensive enhancement by leveraging the spatiotemporal interactions. 

For the implementation with additional visual features of CLIP image encoder, methods with extra features, such as BIG~\cite{gao2022classification}, Social Fabric~\cite{chen2021social}, and Social Fabric-X~\cite{chen2021social}, are also involved for a fair comparison. With the additional features integrated, VrdONE-X achieves the best results for all metrics except for RelTag P@1. Notably, VrdONE-X demonstrates a significant advantage in temporal boundary localization performance, showing improvements of +0.96\%, +0.36\%, and +1.21\% on mAP, R@50, and R@100 of RelDet, respectively. VrdONE balances the tasks between relation classification and temporal localization, maintaining outstanding performance across all tasks. Without simultaneous learning of classification and localization, previous methods exhibit drawbacks in relation detection, especially leading to poor localization results and unsatisfactory performance on the RelDet task. Compared to the ImageNet-VidVRD dataset, the temporal boundaries in VidOR are labeled in finer granularity, and the mean duration of entities is much longer. As a result, more challenges in localizing relations are presented, leading to a more pronounced performance gap between VrdONE and other approaches.

On the ImageNet-VidVRD dataset, VrdONE outperforms the previous state-of-the-art method, HCM~\cite{wei2023defense}, by +1.65\%, +0.23\%, +0.16\%, +2.00\%, +2.00\%, and +0.62\% on all the RelDet and RelTag metrics. By amalgamating the diverse metrics across both datasets, our VrdONE demonstrates exceptional and robust performance on video relation detection, thereby validating the efficacy of spatiotemporal learning and the single-step methodology. 

\subsection{Ablation Studies}
We conduct comprehensive ablation studies on the VidOR dataset to demonstrate the effectiveness of the proposed Bilateral Spatiotemporal Aggregation mechanism and the one-stage query-based 1D temporal relation segmentor. We also evaluate several critical hyper-parameters to affirm the robustness of our method. 

\noindent{\textbf{Subject-Object Synergy Module.}}
Table~\ref{tab:vidor_mcqa_arch_ablation} presents four variants to illustrate the effectiveness of the SOS module, including its absent version (``w/o SOS'') and three different implementations of SOS (``w/o IAB'', ``cross-attention'', and ``IAB''). Without SOS, there is a substantial drop (-0.32\% mAP, -0.37\% R@100, and -0.24\% P@1) on RelDet and RelTag, highlighting the importance of capturing mutual interactions. Our IAB implementation achieves superior spatiotemporal representation perception within entity pairs, showing a notable advantage (+0.26 mAP) compared to the basic model. Notably, our basic model still outperforms previous methods, showcasing the effectiveness of the one-stage learning methodology.

\noindent{\textbf{Number of Queries.}} The number of queries determines the model's video reasoning capability. A tight setting of $N_q$ hinders the modeling of diverse relationships, while excessive queries increase training complexity. BIG~\cite{gao2022classification} leverages a large number of queries (\eg, 192) to detect all relation classes in one video. Conversely, our work independently estimates the relations for each pair, therefore requiring much fewer queries, as demonstrated in Table~\ref{tab:vidor_query_ablation}. This can be extensively supported by a quantitative evaluation indicating that, on average, each subject-object pair in a VidOR single video is associated with only 2.30 relations.

\noindent{\textbf{Number of SOS modules.}}
Table~\ref{tab:vidor_sos_layer_ablation} illustrates the influence of the number of the Subject-Object Synergy layers $L$. We set the number of layers as 2 in practice. 

\noindent{\textbf{Positional Features.}}
In our spatiotemporal synergistic learning, incorporating positional representations $\theta^a$ and $\theta^r$ helps the model perceive the spatial variances of relations. Table~\ref{tab:vidor_spatial_representation_ablation} emphasizes the influences of absolute and relative positional changes on relation detection, indicating the necessity of spatiotemporal perception.

\begin{table}[tp]
    \centering
    \caption{Analysis of the number of the SOS modules $L$.}
    \vspace{-3mm}
    \scalebox{1.0}{
        \begin{tabular}{l|ccc|ccc}
            \toprule
            \multirow{2}{*}{\ $L$} &  \multicolumn{3}{c|}{Relation Detection} & \multicolumn{3}{c}{Relation Tagging} \\
            & mAP & R@50 & R@100 & P@1 & P@5 & P@10 \\
            \midrule
            \ 1 & \underline{11.76} & \textbf{11.16} & \underline{14.18}& \underline{65.87} & 54.30 & 43.52 \\
            \ 2* & \textbf{11.86} & \underline{11.13} & \textbf{14.21} & \textbf{66.11}  & \underline{54.92} & \textbf{43.90} \\
            \ 3 & 11.61 & \underline{11.13} & 14.15 & \underline{65.87}  & \textbf{55.17} & \underline{43.53} \\
            \bottomrule
        \end{tabular}
    }
    \label{tab:vidor_sos_layer_ablation}
    \centering
\end{table}

\begin{table}[tp]
    \centering
    \caption{Ablation on the positional features $\theta^a$ and $\theta^r$.}
    \vspace{-3mm}
    \scalebox{1.0}{
        \begin{tabular}{cc|ccc|ccc}
            \toprule
            \multirow{2}{*}{$\theta^a$} & \multirow{2}{*}{$\theta^r$} & \multicolumn{3}{c|}{Relation Detection} & \multicolumn{3}{c}{Relation Tagging} \\
            & & mAP & R@50 & R@100 & P@1 & P@5 & P@10 \\
            \midrule
            -- & -- & 10.84 & 10.46 & 13.45 & 65.62 & 54.46 & 43.24 \\ 
            -- & \checkmark & 11.45 & 10.75 & 13.85 & \underline{66.11} & 54.36 & \underline{43.82} \\
            \checkmark & -- & \underline{11.83} & \underline{11.06} & \underline{13.98} & \textbf{66.91} & \underline{54.64} & 43.60 \\
            \checkmark & \checkmark & \textbf{11.86} & \textbf{11.13} & \textbf{14.21} & \underline{66.11} & \textbf{54.92} & \textbf{43.90} \\
            \bottomrule
        \end{tabular}
    }
    \label{tab:vidor_spatial_representation_ablation}
    \centering
\end{table}

\subsection{Qualitative Results}
\begin{figure}[t]
    \centering
    \includegraphics[width=\linewidth]{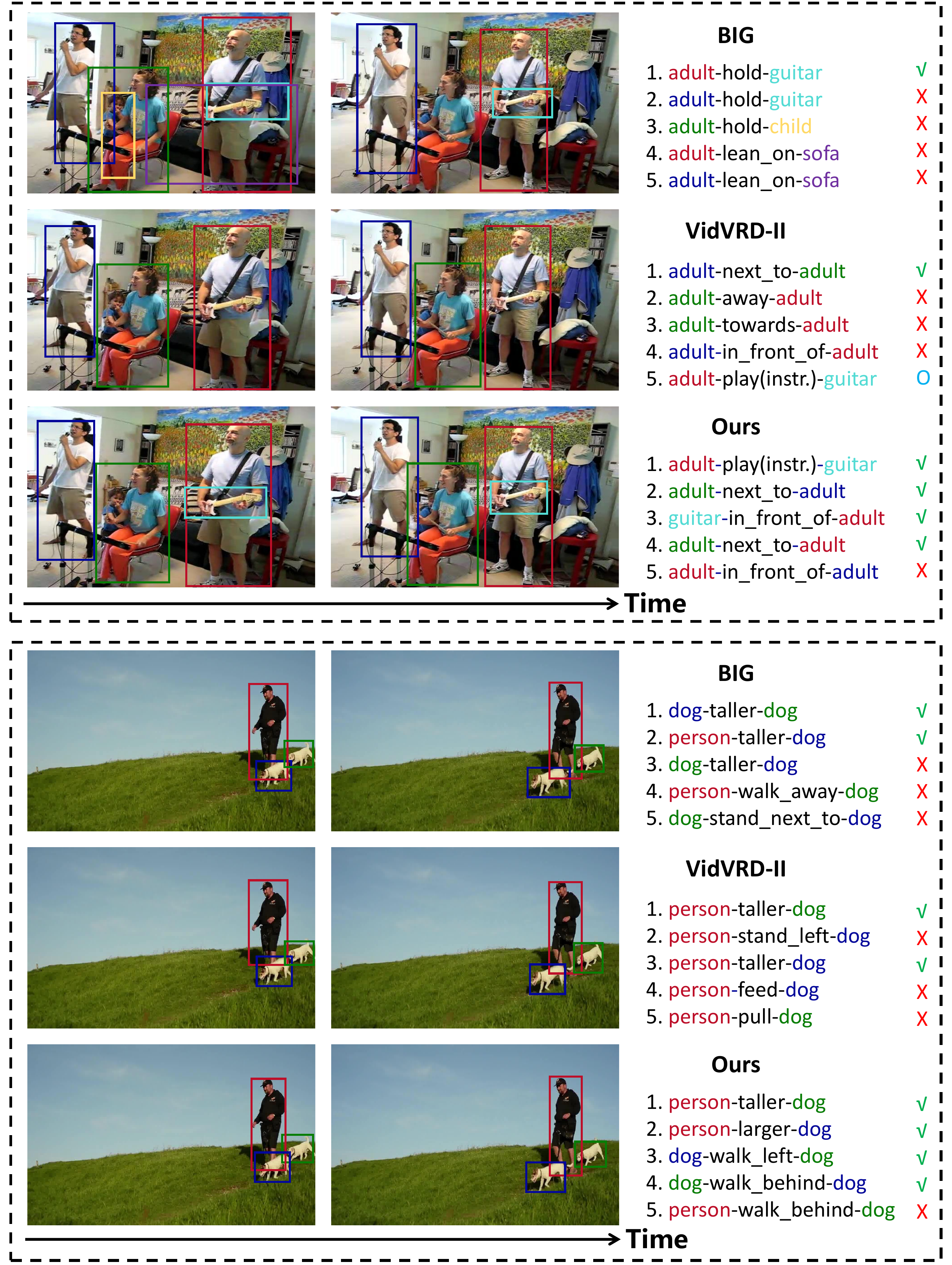}
    \vspace{-6mm}
    \caption{
        Visualization of video relation detection results with open-source methods on the VidOR dataset (top) and ImageNet-VidVRD dataset (bottom). The \textcolor{green}{$\surd$}, \textcolor{red}{$\times$}, and \textcolor{blue}{$\bigcirc$} represent correct, incorrect, and missing detection instances, respectively.
    }
    \label{fig:fig_sec4_exp_visualize}
\end{figure}

In Fig.~\ref{fig:fig_sec4_exp_visualize}, we present several visualization examples comparing our VrdONE with BIG~\cite{gao2022classification} and VidVRD-II~\cite{shang2021video}. The top part of Fig.~\ref{fig:fig_sec4_exp_visualize} exhibits a complex scene that features multiple heavily occluded entities from the VidOR dataset. Our VrdONE precisely captures most of the relations. In contrast, BIG and VidVRD-II suffer from incorrect and missing detections, especially in human-object interactions like ``\textit{\textcolor[RGB]{189,15,41}{adult}-play(instr.)-\textcolor[RGB]{83,219,212}{guitar}}''. In another case drawn from the ImageNet-VidVRD dataset, our VrdONE also produces diverse and confident detection results. It is worth mentioning that VrdONE accurately comprehends size and location relations. For short-lived and enduring relations, such as ``\textit{\textcolor[RGB]{4,8,154}{dog}-walk left-\textcolor[RGB]{8,125,5}{dog}}'' and  ``\textit{\textcolor[RGB]{189,15,41}{person}-taller-\textcolor[RGB]{8,125,5}{dog}}'', VrdONE can discern the spatiotemporal variance patterns between subject-object pairs, affirming its advanced spatiotemporal understanding.

The qualitative experiments demonstrate the superiority of our one-stage method and the effectiveness of spatiotemporal synergistic learning.

\section{Conclusion}
\label{sec:conclusion}

In this paper, we reframe the VidVRD challenge as a 1D instance segmentation problem and unveil VrdONE, a pioneering one-stage detection model designed to improve both predicate classification and localization tasks. The novel Subject-Object Synergy (SOS) module within VrdONE captures mutual interactions between subject-object pairs gradually, thus enhancing video representations. Comprehensive quantitative and qualitative assessments affirm that VrdONE achieves unparalleled performance in its field.

\noindent{\textbf{Limitations.}} Despite VrdONE's advanced capabilities, it does exhibit certain constraints. Its effectiveness is highly impacted by the underlying pretrained video detection and tracking algorithms, as it utilizes processed tracklets for input. Additionally, VrdONE processes all possible subject-object pairs during inference without preliminary filtering, potentially diminishing its overall efficiency.

\begin{acks}
The work is supported by China National Key R\&D Program (Grant No.2023YFE0202700), Key-Area Research and Development Program of Guangzhou City (No.2023B01J0022), Guangdong Provincial Natural Science Foundation for Outstanding Youth Team Project (No.2024B1515040010), National Natural Science Foundation of China (No.62302170).
\end{acks}

\bibliographystyle{ACM-Reference-Format}
\balance
\bibliography{reference}


\begin{thebibliography}{49}


\ifx \showCODEN    \undefined \def \showCODEN     #1{\unskip}     \fi
\ifx \showDOI      \undefined \def \showDOI       #1{#1}\fi
\ifx \showISBNx    \undefined \def \showISBNx     #1{\unskip}     \fi
\ifx \showISBNxiii \undefined \def \showISBNxiii  #1{\unskip}     \fi
\ifx \showISSN     \undefined \def \showISSN      #1{\unskip}     \fi
\ifx \showLCCN     \undefined \def \showLCCN      #1{\unskip}     \fi
\ifx \shownote     \undefined \def \shownote      #1{#1}          \fi
\ifx \showarticletitle \undefined \def \showarticletitle #1{#1}   \fi
\ifx \showURL      \undefined \def \showURL       {\relax}        \fi
\providecommand\bibfield[2]{#2}
\providecommand\bibinfo[2]{#2}
\providecommand\natexlab[1]{#1}
\providecommand\showeprint[2][]{arXiv:#2}

\bibitem[Arnab et~al\mbox{.}(2021)]%
        {arnab2021vivit}
\bibfield{author}{\bibinfo{person}{Anurag Arnab}, \bibinfo{person}{Mostafa Dehghani}, \bibinfo{person}{Georg Heigold}, \bibinfo{person}{Chen Sun}, \bibinfo{person}{Mario Lu{\v{c}}i{\'c}}, {and} \bibinfo{person}{Cordelia Schmid}.} \bibinfo{year}{2021}\natexlab{}.
\newblock \showarticletitle{Vivit: A video vision transformer}. In \bibinfo{booktitle}{\emph{Proceedings of the IEEE International Conference on Computer Vision}}. \bibinfo{pages}{6836--6846}.
\newblock


\bibitem[Ba et~al\mbox{.}(2016)]%
        {ba2016layer}
\bibfield{author}{\bibinfo{person}{Jimmy~Lei Ba}, \bibinfo{person}{Jamie~Ryan Kiros}, {and} \bibinfo{person}{Geoffrey~E Hinton}.} \bibinfo{year}{2016}\natexlab{}.
\newblock \showarticletitle{Layer normalization}.
\newblock \bibinfo{journal}{\emph{arXiv preprint arXiv:1607.06450}} (\bibinfo{year}{2016}).
\newblock


\bibitem[Carion et~al\mbox{.}(2020)]%
        {carion2020end}
\bibfield{author}{\bibinfo{person}{Nicolas Carion}, \bibinfo{person}{Francisco Massa}, \bibinfo{person}{Gabriel Synnaeve}, \bibinfo{person}{Nicolas Usunier}, \bibinfo{person}{Alexander Kirillov}, {and} \bibinfo{person}{Sergey Zagoruyko}.} \bibinfo{year}{2020}\natexlab{}.
\newblock \showarticletitle{End-to-end object detection with transformers}. In \bibinfo{booktitle}{\emph{European Conference on Computer Vision}}. \bibinfo{pages}{213--229}.
\newblock


\bibitem[Carreira and Zisserman(2017)]%
        {carreira2017quo}
\bibfield{author}{\bibinfo{person}{Joao Carreira} {and} \bibinfo{person}{Andrew Zisserman}.} \bibinfo{year}{2017}\natexlab{}.
\newblock \showarticletitle{Quo vadis, action recognition? a new model and the kinetics dataset}. In \bibinfo{booktitle}{\emph{Proceedings of the IEEE Conference on Computer Vision and Pattern Recognition}}. \bibinfo{pages}{6299--6308}.
\newblock


\bibitem[Chen et~al\mbox{.}(2021)]%
        {chen2021social}
\bibfield{author}{\bibinfo{person}{Shuo Chen}, \bibinfo{person}{Zenglin Shi}, \bibinfo{person}{Pascal Mettes}, {and} \bibinfo{person}{Cees~GM Snoek}.} \bibinfo{year}{2021}\natexlab{}.
\newblock \showarticletitle{Social fabric: Tubelet compositions for video relation detection}. In \bibinfo{booktitle}{\emph{Proceedings of the IEEE International Conference on Computer Vision}}. \bibinfo{pages}{13485--13494}.
\newblock


\bibitem[Chen et~al\mbox{.}(2020)]%
        {chen2020memory}
\bibfield{author}{\bibinfo{person}{Yihong Chen}, \bibinfo{person}{Yue Cao}, \bibinfo{person}{Han Hu}, {and} \bibinfo{person}{Liwei Wang}.} \bibinfo{year}{2020}\natexlab{}.
\newblock \showarticletitle{Memory enhanced global-local aggregation for video object detection}. In \bibinfo{booktitle}{\emph{Proceedings of the IEEE Conference on Computer Vision and Pattern Recognition}}. \bibinfo{pages}{10337--10346}.
\newblock


\bibitem[Cheng et~al\mbox{.}(2021)]%
        {cheng2021per}
\bibfield{author}{\bibinfo{person}{Bowen Cheng}, \bibinfo{person}{Alex Schwing}, {and} \bibinfo{person}{Alexander Kirillov}.} \bibinfo{year}{2021}\natexlab{}.
\newblock \showarticletitle{Per-pixel classification is not all you need for semantic segmentation}.
\newblock \bibinfo{journal}{\emph{Advances in Neural Information Processing Systems}} (\bibinfo{year}{2021}), \bibinfo{pages}{17864--17875}.
\newblock


\bibitem[Cheng and Schwing(2022)]%
        {cheng2022xmem}
\bibfield{author}{\bibinfo{person}{Ho~Kei Cheng} {and} \bibinfo{person}{Alexander~G Schwing}.} \bibinfo{year}{2022}\natexlab{}.
\newblock \showarticletitle{Xmem: Long-term video object segmentation with an atkinson-shiffrin memory model}. In \bibinfo{booktitle}{\emph{European Conference on Computer Vision}}. \bibinfo{pages}{640--658}.
\newblock


\bibitem[Choromanski et~al\mbox{.}(2020)]%
        {choromanski2020rethinking}
\bibfield{author}{\bibinfo{person}{Krzysztof Choromanski}, \bibinfo{person}{Valerii Likhosherstov}, \bibinfo{person}{David Dohan}, \bibinfo{person}{Xingyou Song}, \bibinfo{person}{Andreea Gane}, \bibinfo{person}{Tamas Sarlos}, \bibinfo{person}{Peter Hawkins}, \bibinfo{person}{Jared Davis}, \bibinfo{person}{Afroz Mohiuddin}, \bibinfo{person}{Lukasz Kaiser}, {et~al\mbox{.}}} \bibinfo{year}{2020}\natexlab{}.
\newblock \showarticletitle{Rethinking attention with performers}.
\newblock \bibinfo{journal}{\emph{arXiv preprint arXiv:2009.14794}} (\bibinfo{year}{2020}).
\newblock


\bibitem[Danelljan et~al\mbox{.}(2014)]%
        {danelljan2014accurate}
\bibfield{author}{\bibinfo{person}{Martin Danelljan}, \bibinfo{person}{Gustav H{\"a}ger}, \bibinfo{person}{Fahad Khan}, {and} \bibinfo{person}{Michael Felsberg}.} \bibinfo{year}{2014}\natexlab{}.
\newblock \showarticletitle{Accurate scale estimation for robust visual tracking}. In \bibinfo{booktitle}{\emph{British Machine Vision Conference}}.
\newblock


\bibitem[Dosovitskiy et~al\mbox{.}(2020)]%
        {dosovitskiy2020image}
\bibfield{author}{\bibinfo{person}{Alexey Dosovitskiy}, \bibinfo{person}{Lucas Beyer}, \bibinfo{person}{Alexander Kolesnikov}, \bibinfo{person}{Dirk Weissenborn}, \bibinfo{person}{Xiaohua Zhai}, \bibinfo{person}{Thomas Unterthiner}, \bibinfo{person}{Mostafa Dehghani}, \bibinfo{person}{Matthias Minderer}, \bibinfo{person}{Georg Heigold}, \bibinfo{person}{Sylvain Gelly}, {et~al\mbox{.}}} \bibinfo{year}{2020}\natexlab{}.
\newblock \showarticletitle{An image is worth 16x16 words: Transformers for image recognition at scale}.
\newblock \bibinfo{journal}{\emph{arXiv preprint arXiv:2010.11929}} (\bibinfo{year}{2020}).
\newblock


\bibitem[Feichtenhofer et~al\mbox{.}(2019)]%
        {feichtenhofer2019slowfast}
\bibfield{author}{\bibinfo{person}{Christoph Feichtenhofer}, \bibinfo{person}{Haoqi Fan}, \bibinfo{person}{Jitendra Malik}, {and} \bibinfo{person}{Kaiming He}.} \bibinfo{year}{2019}\natexlab{}.
\newblock \showarticletitle{Slowfast networks for video recognition}. In \bibinfo{booktitle}{\emph{Proceedings of the IEEE International Conference on Computer Vision}}. \bibinfo{pages}{6202--6211}.
\newblock


\bibitem[Gao et~al\mbox{.}(2022)]%
        {gao2022classification}
\bibfield{author}{\bibinfo{person}{Kaifeng Gao}, \bibinfo{person}{Long Chen}, \bibinfo{person}{Yulei Niu}, \bibinfo{person}{Jian Shao}, {and} \bibinfo{person}{Jun Xiao}.} \bibinfo{year}{2022}\natexlab{}.
\newblock \showarticletitle{Classification-then-grounding: Reformulating video scene graphs as temporal bipartite graphs}. In \bibinfo{booktitle}{\emph{Proceedings of the IEEE Conference on Computer Vision and Pattern Recognition}}. \bibinfo{pages}{19497--19506}.
\newblock


\bibitem[He et~al\mbox{.}(2022)]%
        {he2022masked}
\bibfield{author}{\bibinfo{person}{Kaiming He}, \bibinfo{person}{Xinlei Chen}, \bibinfo{person}{Saining Xie}, \bibinfo{person}{Yanghao Li}, \bibinfo{person}{Piotr Doll{\'a}r}, {and} \bibinfo{person}{Ross Girshick}.} \bibinfo{year}{2022}\natexlab{}.
\newblock \showarticletitle{Masked autoencoders are scalable vision learners}. In \bibinfo{booktitle}{\emph{Proceedings of the IEEE Conference on Computer Vision and Pattern Recognition}}. \bibinfo{pages}{16000--16009}.
\newblock


\bibitem[He and Lau(2015)]%
        {he2015oriented}
\bibfield{author}{\bibinfo{person}{Shengfeng He} {and} \bibinfo{person}{Rynson~WH Lau}.} \bibinfo{year}{2015}\natexlab{}.
\newblock \showarticletitle{Oriented object proposals}. In \bibinfo{booktitle}{\emph{Proceedings of the IEEE International Conference on Computer Vision}}. \bibinfo{pages}{280--288}.
\newblock


\bibitem[Huang et~al\mbox{.}(2016)]%
        {huang2016stereo}
\bibfield{author}{\bibinfo{person}{Shao Huang}, \bibinfo{person}{Weiqiang Wang}, \bibinfo{person}{Shengfeng He}, {and} \bibinfo{person}{Rynson~WH Lau}.} \bibinfo{year}{2016}\natexlab{}.
\newblock \showarticletitle{Stereo object proposals}.
\newblock \bibinfo{journal}{\emph{IEEE Transactions on Image Processing}} (\bibinfo{year}{2016}), \bibinfo{pages}{671--683}.
\newblock


\bibitem[Huang et~al\mbox{.}(2017)]%
        {huang2017egocentric}
\bibfield{author}{\bibinfo{person}{Shao Huang}, \bibinfo{person}{Weiqiang Wang}, \bibinfo{person}{Shengfeng He}, {and} \bibinfo{person}{Rynson~WH Lau}.} \bibinfo{year}{2017}\natexlab{}.
\newblock \showarticletitle{Egocentric temporal action proposals}.
\newblock \bibinfo{journal}{\emph{IEEE Transactions on Image Processing}} (\bibinfo{year}{2017}), \bibinfo{pages}{764--777}.
\newblock


\bibitem[Larsson et~al\mbox{.}(2016)]%
        {larsson2016fractalnet}
\bibfield{author}{\bibinfo{person}{Gustav Larsson}, \bibinfo{person}{Michael Maire}, {and} \bibinfo{person}{Gregory Shakhnarovich}.} \bibinfo{year}{2016}\natexlab{}.
\newblock \showarticletitle{Fractalnet: Ultra-deep neural networks without residuals}.
\newblock \bibinfo{journal}{\emph{arXiv preprint arXiv:1605.07648}} (\bibinfo{year}{2016}).
\newblock


\bibitem[Li et~al\mbox{.}(2021)]%
        {li2021interventional}
\bibfield{author}{\bibinfo{person}{Yicong Li}, \bibinfo{person}{Xun Yang}, \bibinfo{person}{Xindi Shang}, {and} \bibinfo{person}{Tat-Seng Chua}.} \bibinfo{year}{2021}\natexlab{}.
\newblock \showarticletitle{Interventional video relation detection}. In \bibinfo{booktitle}{\emph{Proceedings of the ACM International Conference on Multimedia}}. \bibinfo{pages}{4091--4099}.
\newblock


\bibitem[Lin et~al\mbox{.}(2023)]%
        {lin2023univtg}
\bibfield{author}{\bibinfo{person}{Kevin~Qinghong Lin}, \bibinfo{person}{Pengchuan Zhang}, \bibinfo{person}{Joya Chen}, \bibinfo{person}{Shraman Pramanick}, \bibinfo{person}{Difei Gao}, \bibinfo{person}{Alex~Jinpeng Wang}, \bibinfo{person}{Rui Yan}, {and} \bibinfo{person}{Mike~Zheng Shou}.} \bibinfo{year}{2023}\natexlab{}.
\newblock \showarticletitle{Univtg: Towards unified video-language temporal grounding}. In \bibinfo{booktitle}{\emph{Proceedings of the IEEE International Conference on Computer Vision}}. \bibinfo{pages}{2794--2804}.
\newblock


\bibitem[Lin et~al\mbox{.}(2017a)]%
        {lin2017feature}
\bibfield{author}{\bibinfo{person}{Tsung-Yi Lin}, \bibinfo{person}{Piotr Doll{\'a}r}, \bibinfo{person}{Ross Girshick}, \bibinfo{person}{Kaiming He}, \bibinfo{person}{Bharath Hariharan}, {and} \bibinfo{person}{Serge Belongie}.} \bibinfo{year}{2017}\natexlab{a}.
\newblock \showarticletitle{Feature pyramid networks for object detection}. In \bibinfo{booktitle}{\emph{Proceedings of the IEEE Conference on Computer Vision and Pattern Recognition}}. \bibinfo{pages}{2117--2125}.
\newblock


\bibitem[Lin et~al\mbox{.}(2017b)]%
        {lin2017focal}
\bibfield{author}{\bibinfo{person}{Tsung-Yi Lin}, \bibinfo{person}{Priya Goyal}, \bibinfo{person}{Ross Girshick}, \bibinfo{person}{Kaiming He}, {and} \bibinfo{person}{Piotr Doll{\'a}r}.} \bibinfo{year}{2017}\natexlab{b}.
\newblock \showarticletitle{Focal loss for dense object detection}. In \bibinfo{booktitle}{\emph{Proceedings of the IEEE International Conference on Computer Vision}}. \bibinfo{pages}{2980--2988}.
\newblock


\bibitem[Liu et~al\mbox{.}(2020)]%
        {liu2020beyond}
\bibfield{author}{\bibinfo{person}{Chenchen Liu}, \bibinfo{person}{Yang Jin}, \bibinfo{person}{Kehan Xu}, \bibinfo{person}{Guoqiang Gong}, {and} \bibinfo{person}{Yadong Mu}.} \bibinfo{year}{2020}\natexlab{}.
\newblock \showarticletitle{Beyond short-term snippet: Video relation detection with spatio-temporal global context}. In \bibinfo{booktitle}{\emph{Proceedings of the IEEE Conference on Computer Vision and Pattern Recognition}}. \bibinfo{pages}{10840--10849}.
\newblock


\bibitem[Liu et~al\mbox{.}(2019)]%
        {liu2019deformable}
\bibfield{author}{\bibinfo{person}{Wenxi Liu}, \bibinfo{person}{Yibing Song}, \bibinfo{person}{Dengsheng Chen}, \bibinfo{person}{Shengfeng He}, \bibinfo{person}{Yuanlong Yu}, \bibinfo{person}{Tao Yan}, \bibinfo{person}{Gehard~P Hancke}, {and} \bibinfo{person}{Rynson~WH Lau}.} \bibinfo{year}{2019}\natexlab{}.
\newblock \showarticletitle{Deformable object tracking with gated fusion}.
\newblock \bibinfo{journal}{\emph{IEEE Transactions on Image Processing}} (\bibinfo{year}{2019}), \bibinfo{pages}{3766--3777}.
\newblock


\bibitem[Liu et~al\mbox{.}(2021)]%
        {liu2021swin}
\bibfield{author}{\bibinfo{person}{Ze Liu}, \bibinfo{person}{Yutong Lin}, \bibinfo{person}{Yue Cao}, \bibinfo{person}{Han Hu}, \bibinfo{person}{Yixuan Wei}, \bibinfo{person}{Zheng Zhang}, \bibinfo{person}{Stephen Lin}, {and} \bibinfo{person}{Baining Guo}.} \bibinfo{year}{2021}\natexlab{}.
\newblock \showarticletitle{Swin transformer: Hierarchical vision transformer using shifted windows}. In \bibinfo{booktitle}{\emph{Proceedings of the IEEE International Conference on Computer Vision}}. \bibinfo{pages}{10012--10022}.
\newblock


\bibitem[Liu et~al\mbox{.}(2022)]%
        {liu2022video}
\bibfield{author}{\bibinfo{person}{Ze Liu}, \bibinfo{person}{Jia Ning}, \bibinfo{person}{Yue Cao}, \bibinfo{person}{Yixuan Wei}, \bibinfo{person}{Zheng Zhang}, \bibinfo{person}{Stephen Lin}, {and} \bibinfo{person}{Han Hu}.} \bibinfo{year}{2022}\natexlab{}.
\newblock \showarticletitle{Video swin transformer}. In \bibinfo{booktitle}{\emph{Proceedings of the IEEE Conference on Computer Vision and Pattern Recognition}}. \bibinfo{pages}{3202--3211}.
\newblock


\bibitem[Loshchilov and Hutter(2017)]%
        {loshchilov2017decoupled}
\bibfield{author}{\bibinfo{person}{Ilya Loshchilov} {and} \bibinfo{person}{Frank Hutter}.} \bibinfo{year}{2017}\natexlab{}.
\newblock \showarticletitle{Decoupled weight decay regularization}.
\newblock \bibinfo{journal}{\emph{arXiv preprint arXiv:1711.05101}} (\bibinfo{year}{2017}).
\newblock


\bibitem[Meinhardt et~al\mbox{.}(2022)]%
        {meinhardt2022trackformer}
\bibfield{author}{\bibinfo{person}{Tim Meinhardt}, \bibinfo{person}{Alexander Kirillov}, \bibinfo{person}{Laura Leal-Taixe}, {and} \bibinfo{person}{Christoph Feichtenhofer}.} \bibinfo{year}{2022}\natexlab{}.
\newblock \showarticletitle{Trackformer: Multi-object tracking with transformers}. In \bibinfo{booktitle}{\emph{Proceedings of the IEEE Conference on Computer Vision and Pattern Recognition}}. \bibinfo{pages}{8844--8854}.
\newblock


\bibitem[Mikolov et~al\mbox{.}(2013)]%
        {mikolov2013efficient}
\bibfield{author}{\bibinfo{person}{Tomas Mikolov}, \bibinfo{person}{Kai Chen}, \bibinfo{person}{Greg Corrado}, {and} \bibinfo{person}{Jeffrey Dean}.} \bibinfo{year}{2013}\natexlab{}.
\newblock \showarticletitle{Efficient estimation of word representations in vector space}.
\newblock \bibinfo{journal}{\emph{arXiv preprint arXiv:1301.3781}} (\bibinfo{year}{2013}).
\newblock


\bibitem[Milletari et~al\mbox{.}(2016)]%
        {milletari2016v}
\bibfield{author}{\bibinfo{person}{Fausto Milletari}, \bibinfo{person}{Nassir Navab}, {and} \bibinfo{person}{Seyed-Ahmad Ahmadi}.} \bibinfo{year}{2016}\natexlab{}.
\newblock \showarticletitle{V-net: Fully convolutional neural networks for volumetric medical image segmentation}. In \bibinfo{booktitle}{\emph{International Conference on 3D Vision}}. \bibinfo{pages}{565--571}.
\newblock


\bibitem[Qian et~al\mbox{.}(2019)]%
        {qian2019video}
\bibfield{author}{\bibinfo{person}{Xufeng Qian}, \bibinfo{person}{Yueting Zhuang}, \bibinfo{person}{Yimeng Li}, \bibinfo{person}{Shaoning Xiao}, \bibinfo{person}{Shiliang Pu}, {and} \bibinfo{person}{Jun Xiao}.} \bibinfo{year}{2019}\natexlab{}.
\newblock \showarticletitle{Video relation detection with spatio-temporal graph}. In \bibinfo{booktitle}{\emph{Proceedings of the ACM International Conference on Multimedia}}. \bibinfo{pages}{84--93}.
\newblock


\bibitem[Radford et~al\mbox{.}(2021)]%
        {radford2021learning}
\bibfield{author}{\bibinfo{person}{Alec Radford}, \bibinfo{person}{Jong~Wook Kim}, \bibinfo{person}{Chris Hallacy}, \bibinfo{person}{Aditya Ramesh}, \bibinfo{person}{Gabriel Goh}, \bibinfo{person}{Sandhini Agarwal}, \bibinfo{person}{Girish Sastry}, \bibinfo{person}{Amanda Askell}, \bibinfo{person}{Pamela Mishkin}, \bibinfo{person}{Jack Clark}, {et~al\mbox{.}}} \bibinfo{year}{2021}\natexlab{}.
\newblock \showarticletitle{Learning transferable visual models from natural language supervision}. In \bibinfo{booktitle}{\emph{International Conference on Machine Learning}}. \bibinfo{pages}{8748--8763}.
\newblock


\bibitem[Ren et~al\mbox{.}(2015)]%
        {ren2015faster}
\bibfield{author}{\bibinfo{person}{Shaoqing Ren}, \bibinfo{person}{Kaiming He}, \bibinfo{person}{Ross Girshick}, {and} \bibinfo{person}{Jian Sun}.} \bibinfo{year}{2015}\natexlab{}.
\newblock \showarticletitle{Faster r-cnn: Towards real-time object detection with region proposal networks}.
\newblock \bibinfo{journal}{\emph{Advances in Neural Information Processing Systems}} (\bibinfo{year}{2015}).
\newblock


\bibitem[Roerdink and Meijster(2000)]%
        {roerdink2000watershed}
\bibfield{author}{\bibinfo{person}{Jos~BTM Roerdink} {and} \bibinfo{person}{Arnold Meijster}.} \bibinfo{year}{2000}\natexlab{}.
\newblock \showarticletitle{The watershed transform: Definitions, algorithms and parallelization strategies}.
\newblock \bibinfo{journal}{\emph{Fundamenta Informaticae}} (\bibinfo{year}{2000}), \bibinfo{pages}{187--228}.
\newblock


\bibitem[Shang et~al\mbox{.}(2019)]%
        {shang2019annotating}
\bibfield{author}{\bibinfo{person}{Xindi Shang}, \bibinfo{person}{Donglin Di}, \bibinfo{person}{Junbin Xiao}, \bibinfo{person}{Yu Cao}, \bibinfo{person}{Xun Yang}, {and} \bibinfo{person}{Tat-Seng Chua}.} \bibinfo{year}{2019}\natexlab{}.
\newblock \showarticletitle{Annotating objects and relations in user-generated videos}. In \bibinfo{booktitle}{\emph{Proceedings of the International Conference on Multimedia Retrieval}}. \bibinfo{pages}{279--287}.
\newblock


\bibitem[Shang et~al\mbox{.}(2021)]%
        {shang2021video}
\bibfield{author}{\bibinfo{person}{Xindi Shang}, \bibinfo{person}{Yicong Li}, \bibinfo{person}{Junbin Xiao}, \bibinfo{person}{Wei Ji}, {and} \bibinfo{person}{Tat-Seng Chua}.} \bibinfo{year}{2021}\natexlab{}.
\newblock \showarticletitle{Video visual relation detection via iterative inference}. In \bibinfo{booktitle}{\emph{Proceedings of the ACM International Conference on Multimedia}}. \bibinfo{pages}{3654--3663}.
\newblock


\bibitem[Shang et~al\mbox{.}(2017)]%
        {shang2017video}
\bibfield{author}{\bibinfo{person}{Xindi Shang}, \bibinfo{person}{Tongwei Ren}, \bibinfo{person}{Jingfan Guo}, \bibinfo{person}{Hanwang Zhang}, {and} \bibinfo{person}{Tat-Seng Chua}.} \bibinfo{year}{2017}\natexlab{}.
\newblock \showarticletitle{Video visual relation detection}. In \bibinfo{booktitle}{\emph{Proceedings of the ACM International Conference on Multimedia}}. \bibinfo{pages}{1300--1308}.
\newblock


\bibitem[Su et~al\mbox{.}(2020)]%
        {su2020video}
\bibfield{author}{\bibinfo{person}{Zixuan Su}, \bibinfo{person}{Xindi Shang}, \bibinfo{person}{Jingjing Chen}, \bibinfo{person}{Yu-Gang Jiang}, \bibinfo{person}{Zhiyong Qiu}, {and} \bibinfo{person}{Tat-Seng Chua}.} \bibinfo{year}{2020}\natexlab{}.
\newblock \showarticletitle{Video relation detection via multiple hypothesis association}. In \bibinfo{booktitle}{\emph{Proceedings of the ACM International Conference on Multimedia}}. \bibinfo{pages}{3127--3135}.
\newblock


\bibitem[Tong et~al\mbox{.}(2022)]%
        {tong2022videomae}
\bibfield{author}{\bibinfo{person}{Zhan Tong}, \bibinfo{person}{Yibing Song}, \bibinfo{person}{Jue Wang}, {and} \bibinfo{person}{Limin Wang}.} \bibinfo{year}{2022}\natexlab{}.
\newblock \showarticletitle{Videomae: Masked autoencoders are data-efficient learners for self-supervised video pre-training}.
\newblock \bibinfo{journal}{\emph{Advances in Neural Information Processing Systems}} (\bibinfo{year}{2022}), \bibinfo{pages}{10078--10093}.
\newblock


\bibitem[Vaswani et~al\mbox{.}(2017)]%
        {vaswani2017attention}
\bibfield{author}{\bibinfo{person}{Ashish Vaswani}, \bibinfo{person}{Noam Shazeer}, \bibinfo{person}{Niki Parmar}, \bibinfo{person}{Jakob Uszkoreit}, \bibinfo{person}{Llion Jones}, \bibinfo{person}{Aidan~N Gomez}, \bibinfo{person}{{\L}ukasz Kaiser}, {and} \bibinfo{person}{Illia Polosukhin}.} \bibinfo{year}{2017}\natexlab{}.
\newblock \showarticletitle{Attention is all you need}.
\newblock \bibinfo{journal}{\emph{Advances in Neural Information Processing Systems}}  \bibinfo{volume}{30} (\bibinfo{year}{2017}).
\newblock


\bibitem[Wang et~al\mbox{.}(2023)]%
        {wang2023videomae}
\bibfield{author}{\bibinfo{person}{Limin Wang}, \bibinfo{person}{Bingkun Huang}, \bibinfo{person}{Zhiyu Zhao}, \bibinfo{person}{Zhan Tong}, \bibinfo{person}{Yinan He}, \bibinfo{person}{Yi Wang}, \bibinfo{person}{Yali Wang}, {and} \bibinfo{person}{Yu Qiao}.} \bibinfo{year}{2023}\natexlab{}.
\newblock \showarticletitle{Videomae v2: Scaling video masked autoencoders with dual masking}. In \bibinfo{booktitle}{\emph{Proceedings of the IEEE Conference on Computer Vision and Pattern Recognition}}. \bibinfo{pages}{14549--14560}.
\newblock


\bibitem[Wang et~al\mbox{.}(2021)]%
        {wang2021actionclip}
\bibfield{author}{\bibinfo{person}{Mengmeng Wang}, \bibinfo{person}{Jiazheng Xing}, {and} \bibinfo{person}{Yong Liu}.} \bibinfo{year}{2021}\natexlab{}.
\newblock \showarticletitle{Actionclip: A new paradigm for video action recognition}.
\newblock \bibinfo{journal}{\emph{arXiv preprint arXiv:2109.08472}} (\bibinfo{year}{2021}).
\newblock


\bibitem[Wei et~al\mbox{.}(2023)]%
        {wei2023defense}
\bibfield{author}{\bibinfo{person}{Meng Wei}, \bibinfo{person}{Long Chen}, \bibinfo{person}{Wei Ji}, \bibinfo{person}{Xiaoyu Yue}, {and} \bibinfo{person}{Roger Zimmermann}.} \bibinfo{year}{2023}\natexlab{}.
\newblock \showarticletitle{In Defense of Clip-based Video Relation Detection}.
\newblock \bibinfo{journal}{\emph{arXiv preprint arXiv:2307.08984}} (\bibinfo{year}{2023}).
\newblock


\bibitem[Wojke et~al\mbox{.}(2017)]%
        {wojke2017simple}
\bibfield{author}{\bibinfo{person}{Nicolai Wojke}, \bibinfo{person}{Alex Bewley}, {and} \bibinfo{person}{Dietrich Paulus}.} \bibinfo{year}{2017}\natexlab{}.
\newblock \showarticletitle{Simple online and realtime tracking with a deep association metric}. In \bibinfo{booktitle}{\emph{IEEE International Conference on Image Processing}}. IEEE, \bibinfo{pages}{3645--3649}.
\newblock


\bibitem[Woo et~al\mbox{.}(2021)]%
        {woo2021and}
\bibfield{author}{\bibinfo{person}{Sangmin Woo}, \bibinfo{person}{Junhyug Noh}, {and} \bibinfo{person}{Kangil Kim}.} \bibinfo{year}{2021}\natexlab{}.
\newblock \showarticletitle{What and when to look?: Temporal span proposal network for video relation detection}.
\newblock \bibinfo{journal}{\emph{arXiv preprint arXiv:2107.07154}} (\bibinfo{year}{2021}).
\newblock


\bibitem[Xu et~al\mbox{.}(2020)]%
        {xu2020transductive}
\bibfield{author}{\bibinfo{person}{Yangyang Xu}, \bibinfo{person}{Chu Han}, \bibinfo{person}{Jing Qin}, \bibinfo{person}{Xuemiao Xu}, \bibinfo{person}{Guoqiang Han}, {and} \bibinfo{person}{Shengfeng He}.} \bibinfo{year}{2020}\natexlab{}.
\newblock \showarticletitle{Transductive zero-shot action recognition via visually connected graph convolutional networks}.
\newblock \bibinfo{journal}{\emph{IEEE Transactions on Neural Networks and Learning Systems}} (\bibinfo{year}{2020}), \bibinfo{pages}{3761--3769}.
\newblock


\bibitem[Yan et~al\mbox{.}(2022)]%
        {yan2022videococa}
\bibfield{author}{\bibinfo{person}{Shen Yan}, \bibinfo{person}{Tao Zhu}, \bibinfo{person}{Zirui Wang}, \bibinfo{person}{Yuan Cao}, \bibinfo{person}{Mi Zhang}, \bibinfo{person}{Soham Ghosh}, \bibinfo{person}{Yonghui Wu}, {and} \bibinfo{person}{Jiahui Yu}.} \bibinfo{year}{2022}\natexlab{}.
\newblock \showarticletitle{VideoCoCa: Video-text modeling with zero-shot transfer from contrastive captioners}.
\newblock \bibinfo{journal}{\emph{arXiv preprint arXiv:2212.04979}} (\bibinfo{year}{2022}).
\newblock


\bibitem[Zhang et~al\mbox{.}(2022)]%
        {zhang2022actionformer}
\bibfield{author}{\bibinfo{person}{Chen-Lin Zhang}, \bibinfo{person}{Jianxin Wu}, {and} \bibinfo{person}{Yin Li}.} \bibinfo{year}{2022}\natexlab{}.
\newblock \showarticletitle{Actionformer: Localizing moments of actions with transformers}. In \bibinfo{booktitle}{\emph{European Conference on Computer Vision}}. \bibinfo{pages}{492--510}.
\newblock


\bibitem[Zheng et~al\mbox{.}(2022)]%
        {zheng2022vrdformer}
\bibfield{author}{\bibinfo{person}{Sipeng Zheng}, \bibinfo{person}{Shizhe Chen}, {and} \bibinfo{person}{Qin Jin}.} \bibinfo{year}{2022}\natexlab{}.
\newblock \showarticletitle{Vrdformer: End-to-end video visual relation detection with transformers}. In \bibinfo{booktitle}{\emph{Proceedings of the IEEE Conference on Computer Vision and Pattern Recognition}}. \bibinfo{pages}{18836--18846}.
\newblock


\end{thebibliography}

\appendix

\section{Overview}
\label{sec:app_overview}

This appendix provides further explanations and results. First, we discuss how to incorporate additional visual features into our model in Sec.~\ref{sec:app_extra_feature}. Next, in Sec.~\ref{sec:app_imple}, we present more implementation details, including how to train our model and detect video relations during inference. Additional ablation studies are outlined in Sec.~\ref{sec:app_exp}. Finally, we present extra visualization examples in Sec.~\ref{sec:app_vis}.

\section{Fusing Additional Features}
\label{sec:app_extra_feature}
Our VrdONE model utilizes a pretrained object detector~\cite{chen2020memory} to extract visual features $F = \{f_1, f_2, ..., f_N \}$ for entities, together with their corresponding spatial positions $\Theta = \{\theta_1, \theta_2, ..., \theta_N \}$, where $\theta_i$ represents the bounding box tubelet for entity $i$. 

In Table~\ref{tab:vidor_metrics}, we integrate additional visual features extracted from the CLIP image encoder~\cite{radford2021learning} into our model. This process is performed before the Bilateral Spatiotemporal Aggregation (BSA) module. By cropping the original images for each frame based on $\theta_i$, we employ the CLIP image encoder to encode the cropped images into feature representations $f^{c}_i$. A simple fusion method is applied using a multilayer perceptron (MLP): 
\begin{align}
    f_i := \mathbf{MLP}(\mathbf{Concat}(f_i, f_i^{c})),
\end{align}
where $:=$ indicates that the original features are replaced by the fused features. These fused features undergo further processing by the BSA module and the one-stage relation detector. Incorporating extra visual features has exhibited further improvements in performance. More sophisticated feature integration designs are left for future work.


\begin{table*}[t]
    \caption{Comparison of oracle detection results. The ``-oracle'' postfix means the model is influenced with ground truth detection trajectories and entities' categories provided.}
    \vspace{-2mm}
    \centering
    \scalebox{0.9}{
        \begin{tabular}{c|cccccc|cccccc}
            \toprule
            \multirow{2}{*}{Method} &  \multicolumn{6}{c|}{Relation Detection} & \multicolumn{6}{c}{Relation Tagging} \\
            & mAP & $\Delta$mAP & R@50 & $\Delta$R@50 & R@100 & $\Delta$R@100 & P@1 & $\Delta$P@1 & P@5 & $\Delta$P@5 & P@10 & $\Delta$P@10 \\
            \midrule
            \multicolumn{13}{c}{ImageNet-VidVRD} \\
            \midrule
            VidVRD~\cite{shang2017video} & 8.58 & & 5.54 & & 6.37 & & 43.00 & & 28.90 & & 20.80 & \\ 
            VidVRD-orcale~\cite{shang2017video} & 15.53 & +6.95 & 12.51 & +6.97 & 16.55 & +10.18 & 43.50 & +0.50 & 29.70 & +0.80 & 23.20 & \textbf{+2.40} \\ 
            \textbf{VrdONE} & 31.33 & & 18.20 & & 21.61 & & 80.50 & & 59.40 & & 44.17 & \\ 
            \textbf{VrdONE-orcale} & 43.15 & \textbf{+11.82} & 30.67 & \textbf{+12.47} & 38.30 & \textbf{+16.69} & 82.50 & \textbf{+2.00} & 62.10 & \textbf{+2.70} & 46.55 & +2.38 \\ 
            \midrule
            \multicolumn{13}{c}{VidOR} \\
            \midrule
            \textbf{VrdONE} & 11.86 & & 11.13 & &14.21 & & 66.11 & & 54.92 & &43.90 &\\ 
            \textbf{VrdONE-orcale} & 41.75 & \textbf{+29.89} & 35.93 & \textbf{+24.80} & 47.79 & \textbf{+33.58} & 85.10 & \textbf{+18.99} & 71.37 & \textbf{+16.45} & 58.43 & \textbf{+14.53} \\ 
            \bottomrule
        \end{tabular}
    }
    \label{tab:app_tracklets_ablation}
    \centering
\end{table*}

\section{More Implementation details}
\label{sec:app_imple}

This section introduces more training and inference details regarding our experiments on the VidOR and ImageNet-VidVRD datasets. 

\subsection{VidOR}

\noindent{\textbf{Parameter Settings.}}
We initialize the feature dimension $C$ to 512. The maximum length of overlapped subject-object durations $l_{so}$ is set to 512. We mask the extra parts for overlapped durations shorter than 512 to prevent attention calculations. The relation encoder $E_{mul}$ comprises 3 blocks, resulting in a downsampling ratio $l_{std}$ of 8 and a feature pyramid with lengths of [512, 256, 128, 64], respectively. The relation decoder $D_{rel}$ consists of 4 layers, with the number of queries $N_q$ set to 9. The feature pyramid is integrated into the temporal mask decoder $D_{msk}$ by lateral connections~\cite{lin2017focal} to recover the original feature length. All attention heads are set to 8. The hidden layer dimension in the Feedforward Network (FFN) subsequent to attention calculation is $4\times$ the input feature dimension. The number of Subject-Object Synergy (SOS) modules $L$ is set to 2, and the window size $k_w$ is set to 9. Since the query tokens are parallel and lack meaningful neighboring relationships, we set the kernel size of $\mathbf{Conv1D}$ for the queries to 1 when computing cross-attention $\mathbf{MCA}$ in the relation decoder $D_{rel}$.

\noindent{\textbf{Training Details.}}
We uniformly sample frames with a stride of 4. When the overlapped length between the subject and object exceeds the maximum length $l_{so}$, we randomly truncate it to 512. Relations with duration more than twice $l_{so}$ are excluded from model training. Additionally, we omit the learning of pairs with more relations than $N_q$. The model is trained for 10 epochs with a batch size of 48. Before local attention and MLP computation, LayerNorm~\cite{ba2016layer} is applied. Drop-out and drop-path~\cite{larsson2016fractalnet} rates are specified as 0 and 0.1. Training VrdONE employs the AdamW optimizer with a learning rate of $2\times10^{-4}$, following a cosine annealing schedule that gradually decreases to $2\times10^{-5}$. Warmup and Exponential Moving Average (EMA) techniques are utilized to enhance and stabilize the training process. No positional encoding is added to the features before attention calculation, consistent with~\cite{zhang2022actionformer}. The loss weights $\lambda_{cls}$, $\lambda_{mf}$, $\lambda_{md}$ are assigned as 2, 2, and 5, respectively.

\begin{table}[!t]
    \caption{Analysis of the window size $k_w$.}
    \vspace{-2mm}
    \centering
    \scalebox{0.9}{
        \begin{tabular}{l|ccc|ccc}
            \toprule
            \multirow{2}{*}{$k_w$} &  \multicolumn{3}{c|}{Relation Detection} & \multicolumn{3}{c}{Relation Tagging} \\
            & mAP & R@50 & R@100 & P@1 & P@5 & P@10 \\
            \midrule
            \ 5 & 11.77 & 11.11 & 14.00 & \textbf{67.31} & 54.84 & \textbf{44.14} \\ 
            \ 7 & \underline{11.83} & 11.04 & \underline{14.14} & 66.11 & \textbf{55.23} & 43.87 \\
            \ 9$^*$ & \textbf{11.86} & \textbf{11.13} & \textbf{14.21} & 66.11 & \underline{54.92} & \underline{43.90} \\
            \ 11 & \textbf{11.86} & \underline{11.12} & 14.05 & \underline{66.59} & 54.75 & 43.42 \\
            \bottomrule
        \end{tabular}
    }
    \label{tab:app_vidor_window_size_ablation}
    \centering
\end{table}

\noindent{\textbf{Inference Details.}}
During inference, we retain entities with mean confidence scores greater than $0.4$ and enumerate the remaining entities to compose all possible subject-object pairs. Pairs with no overlapped duration are discarded. All pairs within a video are divided into two groups based on their lengths: those with lengths not greater than $512$ and the longer pairs. For the former, the same processing steps as during training are applied. For the latter, the $l_{so}$ is set to the maximum length among all pairs in a single video. Following prior studies~\cite{shang2017video,gao2022classification}, we retain the top 6 predicate classification results for each detected instance. For each video, we retain the top 200 results across all detected instances. Foreground probability $>0.5$ in the localization masks indicates where relation instances occur. The temporal boundaries $t_b, t_e$ of the relation instances are determined by the first and last positive mask indices. Further processing of the localization masks is typically unnecessary, as filling holes or applying morphological methods (\eg erosion or dilation) yields minimal improvements.

\subsection{ImageNet-VidVRD}
For ImageNet-VidVRD, some hyper-parameters differ. $l_{so}$ is set to 96, resulting in a feature pyramid with lengths of [96, 48, 24, 12]. The window size $k_w$ is set to 7. We employ a sampling stride of 1, a batch size of 24, and conduct training for 12 epochs. During inference, we retain the top 8 predicate classification results for each detected instance.

\section{Additional Experiments}
\label{sec:app_exp}

In this section, we provide further experimental evidence to explore the potential of our framework under the limited quality of pretrained trackers. We also conduct extra ablations to examine the influence of varying window size $k_w$, maximum training length $l_{so}$, our 1D instance segmentation decoder, and the weight factors $\lambda_{cls}$, $\lambda_{mf}$, and $\lambda_{md}$ used in Eq.~\ref{eq:matching_cost} and Eq.~\ref{eq:mask_cost}.

\noindent{\textbf{Tracklets.}}
Before applying VrdONE, the input videos are pre-processed by a pretrained object tracker, which generates lists of detection bounding boxes for each entity. These bounding boxes $\Theta$ are further used to consolidate spatial context information $\Theta^a$ and $\Theta^r$ into the visual features $F$.

\begin{table}[!t]
    \caption{Analysis of overlapped duration length $l_{so}$.}
    \vspace{-2mm}
    \centering
    \scalebox{0.9}{
        \begin{tabular}{c|ccc|ccc}
            \toprule
            \multirow{2}{*}{\shortstack{Video\\Length}} &  \multicolumn{3}{c|}{Relation Detection} & \multicolumn{3}{c}{Relation Tagging} \\
            & mAP & R@50 & R@100 & P@1 & P@5 & P@10 \\
            \midrule
            \ 256 & \underline{11.72} & 11.04 & 14.01 & \textbf{66.95} & \textbf{55.18} & 43.89 \\ 
            \ 512* & \textbf{11.86} & \textbf{11.13} & \textbf{14.21} & \underline{66.11} & \underline{54.92} & \underline{43.90} \\
            \ 1024 & 11.71 & \underline{11.10} & \underline{14.05} & 65.50 & 54.73 & \textbf{44.17} \\
            \bottomrule
        \end{tabular}
    }
    \label{tab:app_vidor_length_ablation}
    \centering
\end{table}

\begin{table}[!h]
    \caption{Ablation on 1D temporal instance segmentation.}
    \vspace{-2mm}
    \centering
    \scalebox{0.9}{
        \begin{tabular}{l|ccc|ccc}
            \toprule
            \multirow{2}{*}{Approach} &  \multicolumn{3}{c|}{Relation Detection} & \multicolumn{3}{c}{Relation Tagging} \\
            & mAP & R@50 & R@100 & P@1 & P@5 & P@10 \\
            \midrule
            detection & 11.28 & 10.85 & 13.85 & 64.47 & 53.53 & 42.50 \\
            sem. seg. & 11.57 & 10.82 & 13.78 & 66.03 & 54.03 & 43.27 \\
            \midrule
            ins. seg. & \textbf{11.86} & \textbf{11.13} & \textbf{14.21} & \textbf{66.11} & \textbf{54.92} & \textbf{43.90} \\
            \bottomrule
        \end{tabular}
    }
    \label{tab:app_head_arch}
    \centering
\end{table}

We empirically find that the accuracy of the detection tubelets significantly impacts the upper bound of our model's performance. To verify this, we propose providing ground truth tubelets and corresponding class labels for entities in our model. As shown in Table~\ref{tab:app_tracklets_ablation}, after incorporating accurate tracklets, our model shows substantial improvement on both ImageNet-VidVRD and VidOR. On the ImageNet-VidVRD dataset, our model achieves significant gains in 5 out of 6 metrics, surpassing VidVRD~\cite{shang2017video}. This trend is also observed on the VidOR dataset, where more obvious improvements are witnessed, with absolute increases of $29.89\%$, $24.80\%$, $33.58\%$, $18.99\%$, $16.45\%$, and $14.53\%$ on mAP, R@50, R@100, P@1, P@5, and P@10, respectively.

The average temporal length of entities in VidOR is much longer than in ImageNet-VidVRD, making the tracklet detection more challenging for trackers and leading to reduced performance. However, even in such a challenging scenario, our model can still perform precise relation classification and temporal localization when supported by accurate tracklets. These results underscore the potential of our proposed VrdONE framework,  particularly when enhanced by advancements in modern object trackers.

\noindent{\textbf{Window Size of Local Attention.}} Table~\ref{tab:app_vidor_window_size_ablation} examines the influence of varying the window size $k_w$. In all our experiments, we select $k_w$ as 9 for balanced performances on RelDet and RelTag.

\noindent{\textbf{Video length.}} Table~\ref{tab:app_vidor_length_ablation} demonstrates the impact of the length of input overlapped subject-object pairs. Empirically, we truncate/pad the videos to a uniform length of 512 for optimal performance.

\noindent{\textbf{Decoder Architecture.}} 
Table~\ref{tab:app_head_arch} illustrates the advantages of our proposed 1D temporal instance segmentation paradigm. We designed two variants by replacing the instance mask generation decoder (``ins. seg.'') with a boundary regression decoder (``detection'') and a semantic mask generation decoder (``sem. seg.''), respectively. Our instance segmentation paradigm achieves the best performance on both RelDet and RelTag.

\begin{figure*}[tp]
    \subfloat[]{
        \includegraphics[width=0.3\linewidth]{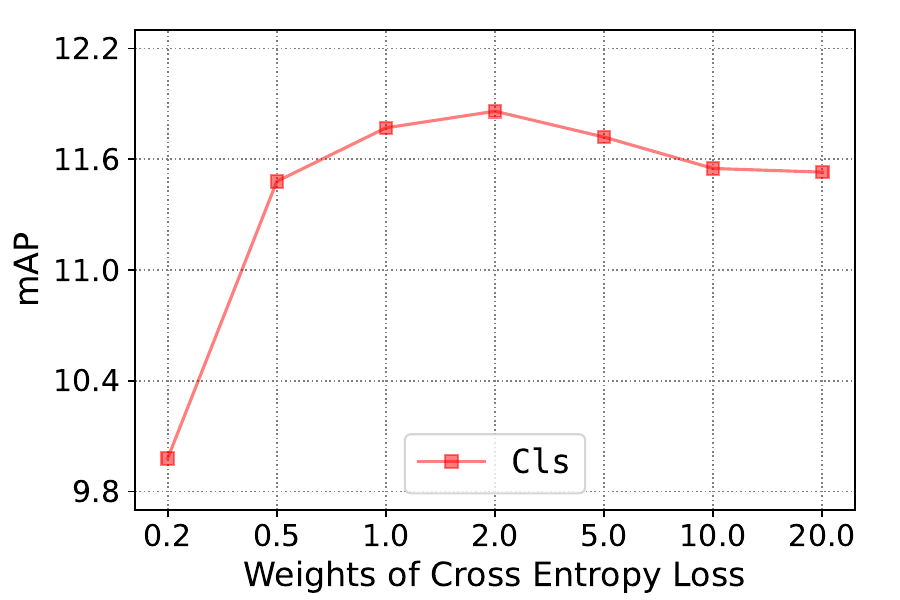}
    }
    \subfloat[]{
        \includegraphics[width=0.3\linewidth]{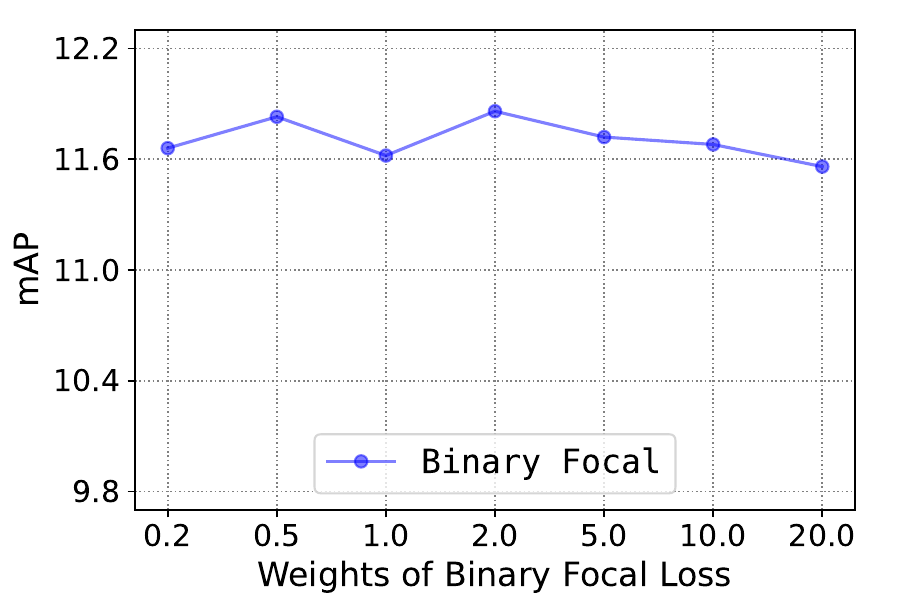}
    }
    \subfloat[]{
        \includegraphics[width=0.3\linewidth]{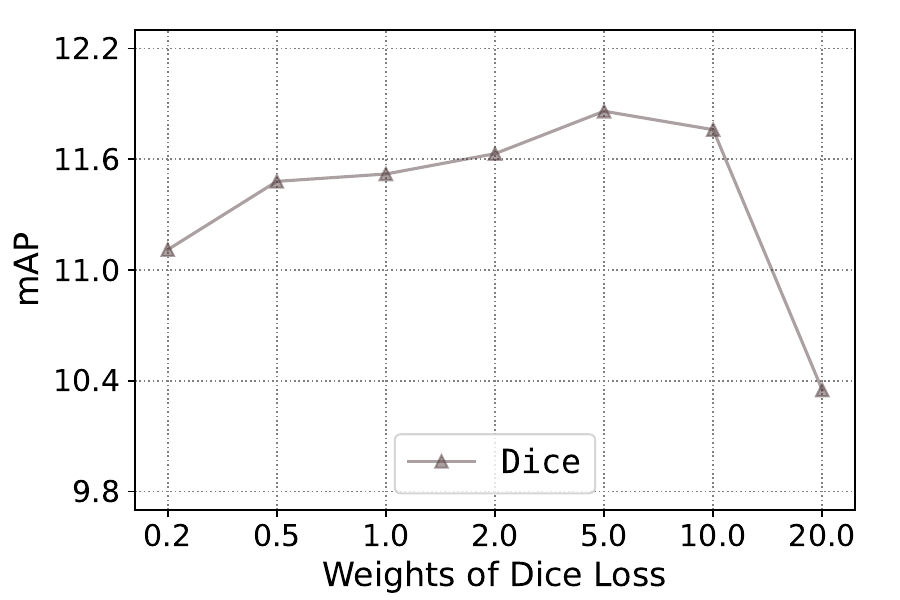}
    }
    \vspace{-3mm}
    \caption{
        Analysis of varying weight factors $\lambda_{cls}$, $\lambda_{mf}$, and $\lambda_{md}$, which are applied to (a) Cross Entropy Loss for relation classification, (b) Mask Focal Loss and (c) Dice Loss for relation localization, respectively.
    }
    \vspace{-3mm}
    \label{fig:app_vidor_loss_factor_ablation}
\end{figure*}

\begin{figure*}[tp]
    \centering
    \scalebox{1.0}{
        \subfloat[]{
            \includegraphics[width=\linewidth]{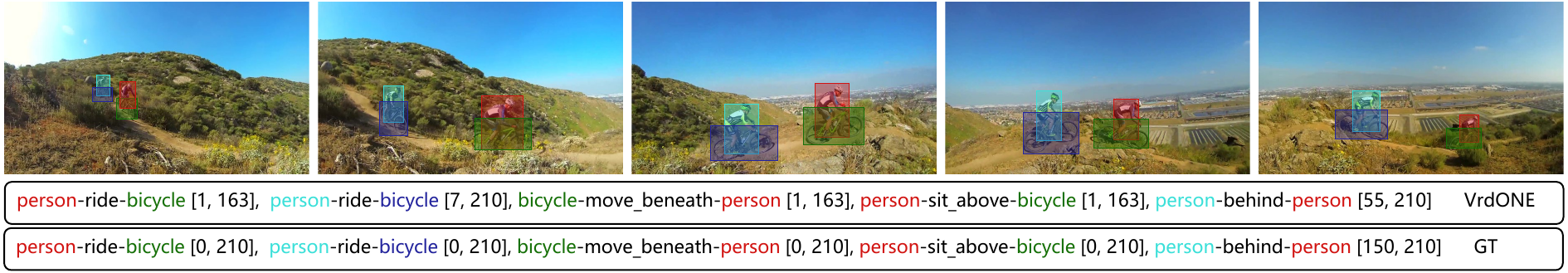}
        }
    }

    \scalebox{1.0}{
        \subfloat[]{
            \includegraphics[width=\linewidth]{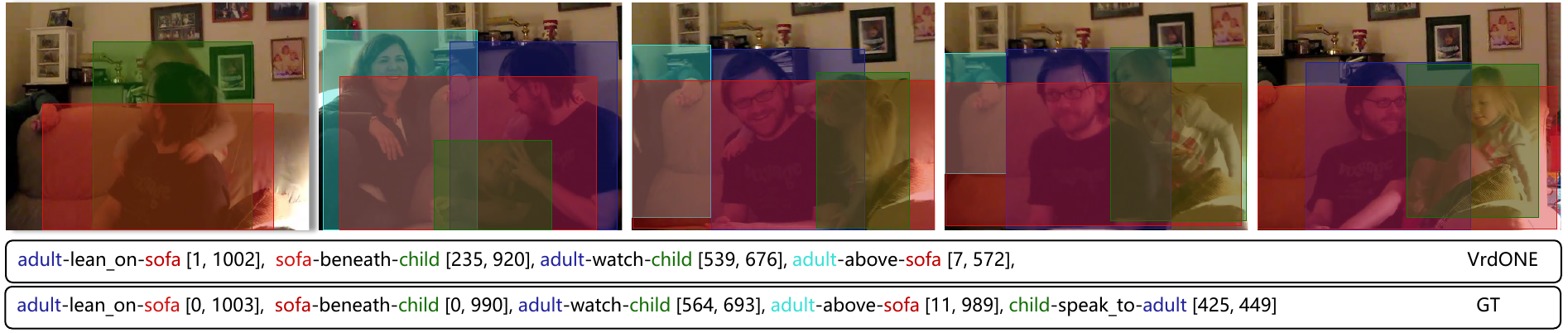}
        }
    }
    \vspace{-3mm}
    \caption{
        Additional qualitative results on ImageNet-VidVRD (above) and VidOR (below). The results are visualized in spatiotemporal forms to include all elements of video relations. (\Eg, subject/object class, subject/object trajectory, predicate class, predicate temporal boundary.) The numbers in the brackets are the start frames and end frames of current relations.
    }
    \vspace{-3mm}
    \label{fig:app_fig_vis1}
\end{figure*}

\noindent{\textbf{Loss Factors.}}
In all our experiments, we set the weight factors $\lambda_{cls}$, $\lambda_{mf}$, and $\lambda_{md}$ as 2, 2, and 5, respectively, based on the performance of mAP. We analyze these factors by changing one of them while keeping the others fixed. The results, depicted in Fig.~\ref{fig:app_vidor_loss_factor_ablation}, demonstrate that our model's performance remains robust within a reasonable range of these factors. However, excessively small $\lambda_{cls}$ or overly large $\lambda_{md}$ leads to drastic drops in performance.

\section{Visualization}
\label{sec:app_vis}

\begin{figure*}[tp]
    \centering
    \scalebox{1.0}{
        \subfloat[]{
            \includegraphics[width=\linewidth]{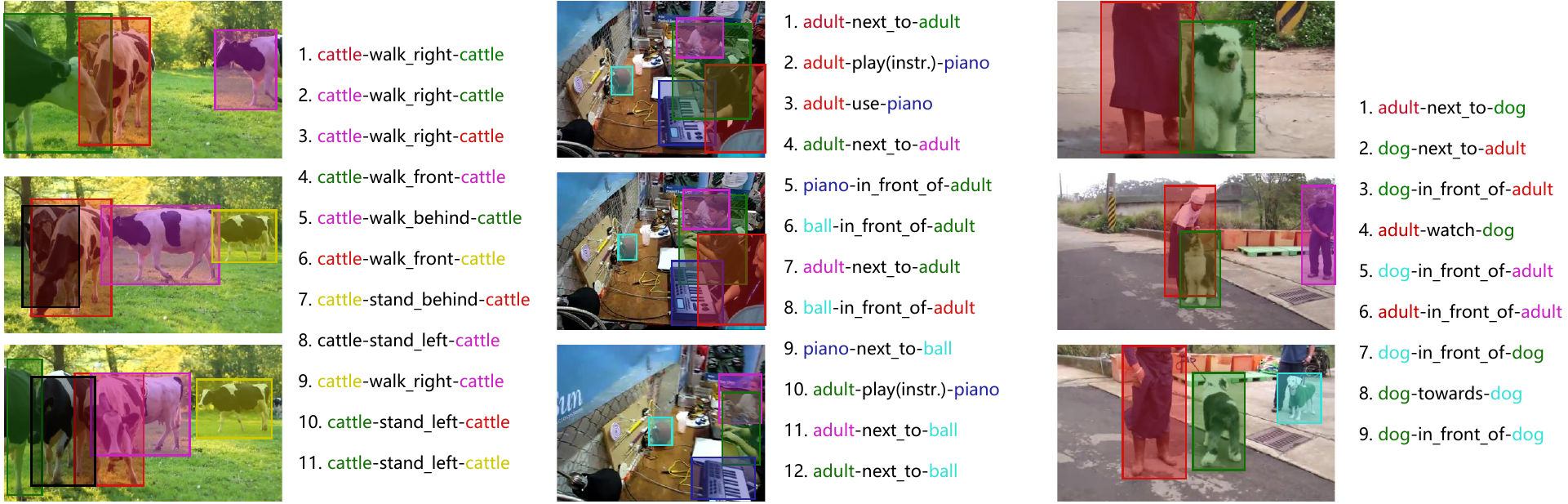}
        }
    }
    \vspace{-3mm}
    \caption{
        Visualizations of the relation classification results in more details. 
    }
    \label{fig:app_fig_vis2}
\end{figure*}

Additional qualitative results are illustrated in Fig.~\ref{fig:app_fig_vis1} and Fig.~\ref{fig:app_fig_vis2}. We present the spatiotemporal visualization of the detection results for both ImageNet-VidVRD (Fig.~\ref{fig:app_fig_vis1}(a)) and VidOR (Fig.~\ref{fig:app_fig_vis1}(b)) datasets. Our method effectively learns relation categories, especially those involving spatial positions, demonstrating the effectiveness of our spatiotemporal learning approach. However, it is evident that the performance of our method is significantly affected by the quality of the tracklets. For instance, in Fig.~\ref{fig:app_fig_vis1}(a), our method accurately identifies that the \rstred{preson} is ridding the \rstgreen{bicycle} across the entire overlapped duration of \rstred{preson}-\rstgreen{bicycle} pair. Nonetheless, the temporally overlapping trajectories are limited to the range [1, 163], decreasing the accuracy of temporal boundary predictions. Similarly, the relation ``\rstred{sofa}-beneath-\rstgreen{child}'' in Fig.~\ref{fig:app_fig_vis1}(b) also highlights this issue. In Fig.~\ref{fig:app_fig_vis2}, we provide additional classification results from our model. Although the query number per pair is limited, the accumulated detection results are sufficient for a comprehensive analysis of the entire video.

\end{document}